\documentclass[10pt,twocolumn,letterpaper]{article}

\usepackage{wacv}              

\usepackage{graphicx}
\usepackage{amsmath}
\usepackage{amssymb}
\usepackage{booktabs}
\usepackage{indentfirst}
\usepackage[
  pagebackref,
  breaklinks,
  colorlinks,
  citecolor=blue,
  linkcolor=blue,   
]{hyperref}
\usepackage{array}
\usepackage{multirow}
\usepackage[table]{xcolor}
\usepackage{pifont,booktabs}
\usepackage{makecell}
\usepackage{float}

\usepackage{algorithm}
\usepackage{algorithmic}

\usepackage[pagebackref,breaklinks,colorlinks]{hyperref}

\usepackage[capitalize]{cleveref}
\crefname{section}{Sec.}{Secs.}
\Crefname{section}{Section}{Sections}
\Crefname{table}{Table}{Tables}
\crefname{table}{Tab.}{Tabs.}

\def\wacvPaperID{1230} 
\def\confName{WACV}
\def\confYear{2025}

\begin{document}

\title{DRWKV: Focusing on Object Edges for Low-Light Image Enhancement}

\author{
Xuecheng Bai\textsuperscript{1}\footnotemark[1]\quad \quad
Yuxiang Wang\textsuperscript{2}\footnotemark[1]\quad \quad
Boyu Hu\textsuperscript{3}\quad \quad
Qinyuan Jie\textsuperscript{1}\\
Chuanzhi Xu\textsuperscript{2}\footnotemark[2]\quad \quad
Kechen Li\textsuperscript{4}\quad \quad
Hongru Xiao\textsuperscript{5}\quad \quad
Vera Chung\textsuperscript{2}
}

\makeatletter
\def\ps@accepted{%
  \def\@oddhead{%
    \hfil
    \large\textcolor{gray}{This paper has been accepted to WACV 2026}%
    \hfil
  }%
  \def\@evenhead{\@oddhead}%
  \def\@oddfoot{}%
  \def\@evenfoot{}%
}
\makeatother

\maketitle

\begingroup
\renewcommand\thefootnote{\fnsymbol{footnote}}
\footnotetext[1]{Equal contribution.}
\footnotetext[2]{Corresponding author: Chuanzhi Xu (chuanzhi.xu@sydney.edu.au).\\
\textbf{Affiliations:}\\
\textsuperscript{1}Shenyang Ligong University, Shenyang, China \\
\textsuperscript{2}The University of Sydney, NSW, Australia \\
\textsuperscript{3}University of International Business and Economics, Beijing, China \\
\textsuperscript{4}Nanjing University of Aeronautics and Astronautics, Nanjing, China\\
\textsuperscript{5}Tongji University, Shanghai, China }
\endgroup
\begin{abstract}
\thispagestyle{accepted}
Low-Light Image Enhancement (LLIE) remains a challenging task, particularly in preserving object edge continuity and fine structural details under extreme illumination degradation. In this paper, we propose a novel model, DRWKV (Detailed Receptance Weighted Key Value), which integrates our proposed Global Edge Retinex (GER) theory, enabling effective decoupling of illumination and edge structures for enhanced edge fidelity. Secondly, we introduce Evolving WKV Attention, a spiral-scanning mechanism that captures spatial edge continuity and models irregular structures more effectively. Thirdly, we design the Bilateral Spectrum Aligner Block (Bi-SAB) and a tailored MS$^{2}$-Loss to jointly align luminance and chrominance features, improving visual naturalness and mitigating artifacts. Extensive experiments on five LLIE benchmarks demonstrate that DRWKV achieves leading performance in PSNR, SSIM, and NIQE while maintaining low computational complexity. Furthermore, DRWKV enhances downstream performance in low-light multi-object tracking tasks, validating its generalization capabilities. The code are available at: \textcolor{blue}{https://github.com/JackBaixue/DRWKV}

\end{abstract}

\section{Introduction}
\label{sec:intro}
Low-Light Image Enhancement (LLIE) technology is regarded as an active method to brighten low-light images at low cost~\cite{nguyen2024diffusion,Tan_2024_ACCV}. Recently, the academic community has primarily explored prominent methods such as global attention modeling~\cite{xu2025upt,wang2024low,shang2024holistic}, local feature enhancement~\cite{zhang2024cnn,guo2016lime,guo2020zero}, and selective state updating~\cite{zou2024wave,qu2024double}. Based on the robust Retinex theory~\cite{8304597}, the composition of a low-light image $I$ can be interpreted as:
\begin{equation}
I = R \cdot L + N,
\end{equation}
where $R$ denotes scene reflectance, $L$ represents the illumination component, and $N$ stands for intrinsic noise.
\begin{figure}[t]
    \centering
    \includegraphics[width=\linewidth]{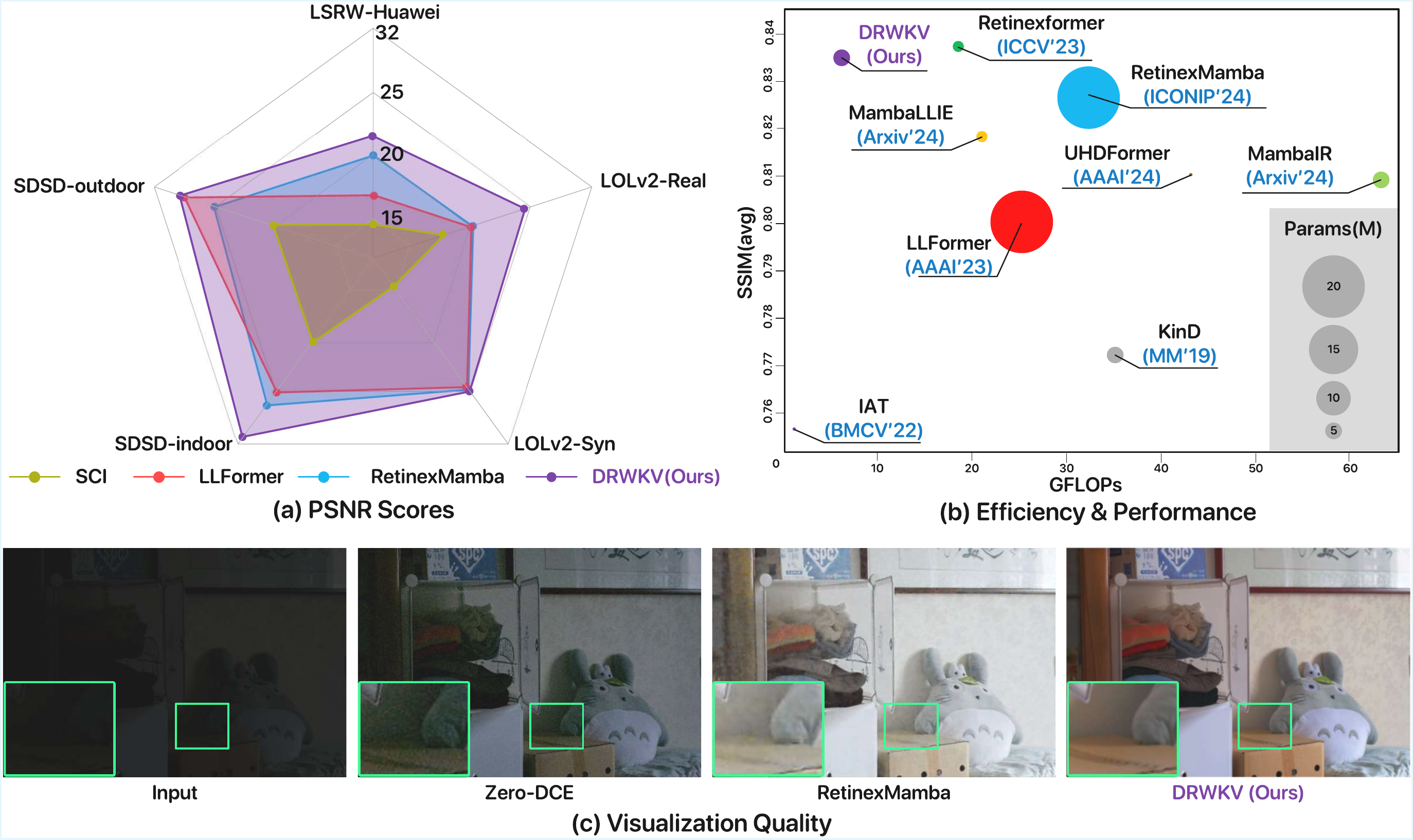}
    \hypertarget{fig:intro-a}{}
    \hypertarget{fig:intro-b}{}
    \hypertarget{fig:intro-c}{}
    \caption{Comparison of our DRWKV with different methods: (a) DRWKV achieves the best PSNR in benchmark tests; (b) DRWKV balances efficiency, parameters, and computational load; (c) DRWKV exhibits more robust performance in edge details.}
    \label{fig:intro}
    \vspace{-1em}
\end{figure}

When constructing a low-light scene, for the image, in addition to the increase in noise \( N \), the illumination component \( L \) is decomposed into global and local illumination. The global spectrum and contour light field will dynamically change under light reflection. In the context of low-light image quality improvement, focusing solely on global illumination will amplify background noise~\cite{shi2024zero} and cause edge distortion~\cite{wang2024multimodal}, while focusing only on contours in isolation leads to rigid edge transitions. Therefore, under the condition that global exposure is controllable, the essence of low-light image enhancement lies in the refinement of object edges. As shown in Figure~\ref{fig:intro}\hyperlink{fig:intro-c}{(c)}, ZeroDCE~\cite{guo2020zero} suffers from edge distortion, and RetinexMamba~\cite{bai2024retinexmamba} exhibits detail blurring, indicating that current methods have limitations in handling extremely low-light environments.

To address these issues, we propose \textbf{\underline{D}etailed \underline{R}eceptance \underline{W}eighted \underline{K}ey \underline{V}alue (\underline{DRWKV})} model, which optimizes the Bi-WKV Attention of VRWKV (in ICLR 2025)~\cite{duan2024vision}—a classic model in the visual RWKV domain. We introduce the \textbf{Evolving WKV Attention}, a novel attention mechanism with superior spatial topological awareness, which can maximize the capture of continuous spatial information of object edges, thereby strengthening the modeling effect of irregular spatial structures. Following this, we design an auxiliary technique—\textbf{Bilateral Spectrum Aligner Block (Bi-SAB)}. By aligning brightness and color spectral features, Bi-SAB resolves issues of over-exposure and color distortion caused by separate processing. Coupled with \textbf{MS²-Loss}, Bi-SAB renders the illumination and colors of enhanced images more natural. We conducted extensive image-enhancement and low-light object-tracking experiments on five benchmark datasets. As illustrated in Figure~\ref{fig:intro}\hyperlink{fig:intro-a}{(a)} and~\ref{fig:intro}\hyperlink{fig:intro-b}{(b)}, our method achieves leading PSNR, SSIM, and NIQE scores, demonstrating significant superiority in edge continuity and detail fidelity.

Our contributions can be summarized as follows:
\begin{itemize}
  \item We propose \textbf{Detailed Receptance Weighted Key Value (DRWKV)} model, for the first time, integrating our proposed \textbf{Global Edge Retinex (GER)} theory with the VRWKV model. It effectively enhances the edge of the low-light image.
  \item We design a novel \textbf{Evolving Scanning (ES) mechanism} applied to WKV Attention, also namely \textbf{Evolving WKV Attention}. It mines edge details from the inside out, enhancing DRWKV's in modeling irregular spatial structures of edges. 
  \item We introduce \textbf{Bilateral Spectrum Aligner Block (Bi-SAB)}, which enriches the color domain of brightened low-light images by coordinating spectral features and coupling with our proposed \textbf{MS²-Loss}.
\end{itemize}
\section{Related Work}
\subsection{Object Edge Perception and Extraction}
Edge perception and extraction techniques are built upon traditional methods~\cite{chen2024highly,lu2022low,9328179,lim2020dslr} and deep learning approaches~\cite{wang2022eanet,guo2020zero}. They achieve fine-grained restoration of object edges in low-light scenarios by fusing decoupled edge features with contextual information.

Different from traditional CNN-based methods such as Retinex-Net~\cite{wei2018deep} and LLNet~\cite{lore2017llnet} that rely on local convolutions to extract edge features, emerging edge-aware approaches introduce topological modeling~\cite{zhao2024dyedgegat,li2023edge,tang2025spiralmamba}. These methods treat edges as ``spatial sequences'' and capture cross-regional features through state accumulation. This design offers a core advantage in enhancing weak edge signals by achieving balanced transitions and local optimization through the adjustment of edge gradient consistency constraints and dynamic thresholds.

\begin{figure*}[t]
    \centering
    \includegraphics[width=\linewidth]{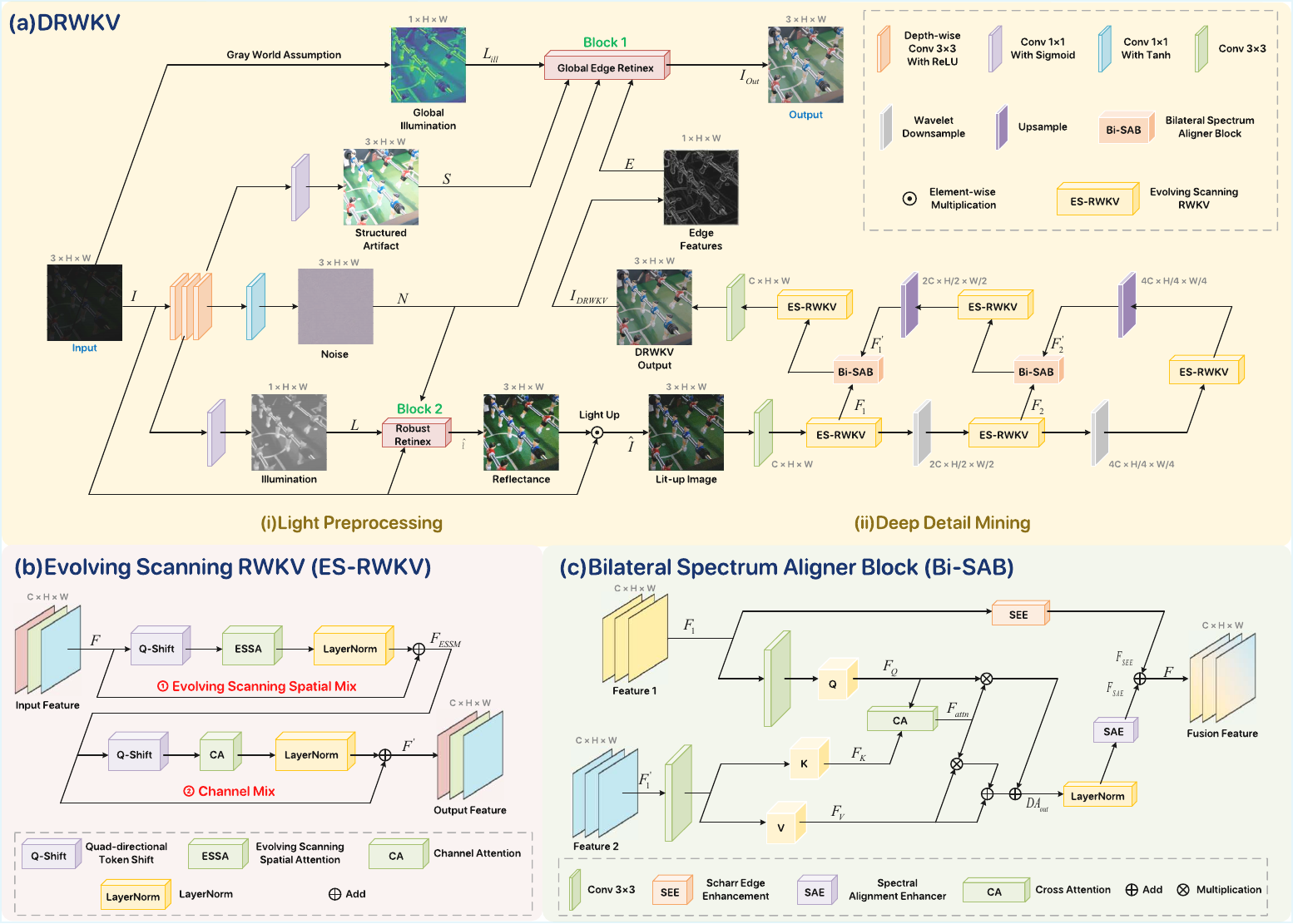}
    \caption{The overview of the DRWKV model. (a) DRWKV processes low-light images through two steps: (i) Light Preprocessing and (ii) Deep Detail Mining. The positions of Block1 and Block2 correspond to the layout settings of the two Retinex theories. (b) The block structure of Evolving Scanning RWKV, which embeds two core mixing modules: Evolving Scanning Spatial Mix and Channel Mix. (c) The block structure of the Bilateral Spectrum Aligner Block.}
    \label{fig:newtork}
\end{figure*}

\subsection{Low-Light Object Edge Restoration and Enhancement}
Previous efforts to improve object edge accuracy in complex low-light environments primarily focus on two directions: exploration of diverse model architectures and optimization of edge enhancement algorithms.

Techniques such as EDTER~\cite{pu2022edter} and Retinex-RAWMamba~\cite{chen2024retinex} explore diverse architectures by leveraging self-attention to capture cross-regional dependencies and linear scanning to optimize temporal accumulation.

Robust edge extraction is achieved through frequency-domain feature filtering~\cite{guan2021dual}, multi-modal edge complementarity~\cite{qian2021improved,liu2024optical}, and adversarial learning-based edge constraints~\cite{bai2024retinexmamba}, further precisely enhancing edge features.

Despite these advances, the modeling of object edge consistency under the visual RWKV framework has received limited attention, particularly in complex low-light scenarios. This suggests a promising direction for integrating edge-aware fidelity modeling into RWKV-based architectures.
\section{Methodology}
\noindent\textbf{Rethinking Edge Fidelity Modeling in VRWKV:} The core of edge fidelity lies in strong consistency between structure and features. The limitations of VRWKV in low-light scenarios are analyzed from two aspects:

\noindent\underline{\textit{(1) Neural Architecture.}} VRWKV adopts a single recursive structure, which mismatches the hierarchical features of edges. This causes aliasing of shallow and deep features, manifested as the failure to decouple illumination variations from edge representation, and leads to frequent misjudgment between noise and edges in low-light environments.

\noindent\underline{\textit{(2) Computational Geometry.}} Edges are low-dimensional manifolds in high-dimensional pixel spaces, becoming more curved under low light due to drastic illumination fluctuations. However, VRWKV assumes features are uniformly distributed in Euclidean space, while edge manifolds actually reside in curved Riemannian spaces. This results in inaccurate Euclidean distance metrics (causing misclassification of edge features) and mismatching between the translation invariance of convolutions and edge scale variations under low light.

\noindent\textbf{Based on the results of our reasoning, we propose two hypotheses:} (1) Hierarchical detail enhancement depends on a gradient-guided mechanism with illumination invariance. (2) Modeling the continuity of low-light edges requires a spatiotemporal alignment modeling scheme compatible with the topological properties of edges.

\noindent\textbf{DRWKV Model:} Different from current mainstream methods, model adopts a layered optimization strategy, which prioritizes combining the gray-world assumption with a three-branch channel consisting of noise $N$, artifact $S$, and illumination $L$ to implement (\ref{Light-Pre}) Light Preprocessing (LP), thereby obtaining the brightness-restored image $\hat{I}$~\cite{cai2023retinexformer}. For the Deep Detail Mining (DDM) process, the RWKV structure comprises a set formed by four groups of ES-RWKV and sampling blocks. For the encoder, (\ref{sec3.2}) Evolving Scanning mechanism is applied for the first time, combined with wavelet transform downsampling to accomplish edge gradient feature extraction. The decoder integrates the sampling information $F$ and $F'$ via the (\ref{BSA}) Bilateral Spectrum Aligner Block (Bi-SAB), enabling progressive enhancement from microscopic textures to macroscopic contours. The overall structure of our DRWKV model is provided in Figure~\ref{fig:newtork}.

\subsection{Light Preprocessing}
\label{Light-Pre}

The Light Preprocessing (LP) structure employs a dual-path parallel guidance approach, balancing brightness, preserving details, and ensuring the baseline quality of LLIE results for the subsequent Deep Detail Mining (DDM) process. Technically, we recalculate the RGB components using the gray-world assumption \( L_{ill} \):
\begin{equation}L_{\text{ill}} = \begin{bmatrix} R' \ G' \ B' \end{bmatrix} = \text{diag}\left( \frac{\frac{1}{3} \sum_{i=1}^{N} \left| c_i \right|}{\sum_{i=1}^{N} c_i} \right) \cdot c,\end{equation}
where \( c_i = [R_i, G_i, B_i]^T \) restructures the three-channel vector of a pixel. The detailed procedure involves converting the vector into a diagonal matrix via \( \text{diag}(\cdot) \) and then computing the inner product with the original channel vector \( c \). By combining \( L_{ill} \) with the artifact component \( S \), we explicitly define the interference boundaries to be avoided, enabling the Global Edge Retinex (GER) to enhance brightness without amplifying artifacts, thus achieving the first path of clean baseline enhancement.

We recognize that separating image information into an illumination component \( L \) and a noise component \( N \) allows for a more refined representation:

\begin{equation}
\hat{I} = I \times \hat{R} = I \times (I - N)/L.
\end{equation}

In the second path, we process different information components by computing the product of the reconstructed reflectance component \( \hat{R} \) and the input low-light image \( I \), resulting in a low-light enhanced image \( \hat{I} \) that retains the original detail information.
lts for the subsequent RWKV stage.

\subsection{Deep Detail Mining} \label{sec3.2}

\noindent{\textbf{Global Edge Retinex Theory}}

The traditional Robust Retinex theory has limitations in component decoupling and deep/shallow detail processing. Based on the hierarchical detail enhancement approach in \textit{Hypothesis(1)}, we propose the Global Edge Retinex (GER) theory, and the core formula is as follows:

\begin{equation}
I = (R + \alpha\cdot E) \odot L_{ill} + \beta\cdot N + \gamma\cdot S.
\end{equation}
The theory takes the explicit edge feature \( E \) (the formula is provided in Appendix \textcolor{blue}{3.3}), the spatially heterogeneous noise \( N \) (obtained after segmenting traditional noise), the gray-world assumption \( L_{ill} \), the artifact \( S \), and the learnable parameters \( \alpha \), \( \beta \), \( \gamma \) (which control feature expression) as the anchors for addressing low-light edge detail issues.

The specific breakdown is as follows: Deep edge structures exhibit illumination invariance, so GER uses the reflectance component \( R \) as the stable base:
\begin{equation}
R = \frac{I - N}{L_{ill} + \epsilon} \in [0, 1],
\end{equation}
where \( R \) strips off spatially heterogeneous noise via \( I - N \) and neutralizes illumination interference through \( L_{ill} + \epsilon \), providing a "clean" benchmark for detail enhancement.

From the model perspective, the edge feature \( E \) (processed by spiral gradient scanning) is used to construct the enhanced reflectance map:
\begin{equation}
R_{\text{enh}} = R + \alpha \cdot E.
\end{equation}
The parameter \( \alpha \) is used to control the intensity of differential enhancement of deep/shallow details guided by \( E \), ensuring clear feature hierarchy. Meanwhile, gradient information is incorporated to enhance edge continuity.

Finally, an enhanced image with edge fidelity is obtained through the collaborative balance of all components:
\begin{equation}
\hat{I} = R_{\text{enh}} \times L_{ill} + \beta \cdot N + \gamma \cdot S.
\end{equation}

\noindent{\textbf{Evolving Scanning Mechanism}}

Evolving Scanning (ES) mechanism addresses \textit{Hypothesis(2)} by transforming the geometric continuity of edge manifolds into temporal continuity capturable by the VRWKV propagation mechanism. Its mathematical foundation lies in the Archimedean spiral equation:  
\begin{equation}
P(\theta) = \left( r(\theta)\cos\theta, \ r(\theta)\sin\theta \right),
\end{equation}
where \(r(\theta) = a + b\theta\) governs the initial radius (\(a\)) and expansion rate (\(b\)) of the spiral. The path expands uniformly from the starting point, naturally adapting to the edge characteristics of spreading from local to global regions. In terms of the scanning mechanism, we extend a single spiral to a four-directional rectangular spiral system. Let \( P \) denote the spiral path spreading outward from the center, and let \( P_i = (x_i, y_i) \) represent the coordinates of the \( i \)-th step (where \( i \in \mathbb{N} \)). The coordinates satisfy:

\begin{equation}
\begin{split}
P_i &= (x_0, y_0) + \left(r_0 + \alpha \lfloor i/4 \rfloor\right) \cdot \delta_{imod4}, \\
    &\quad \text{where } \delta \in \{(0, -1), (1, 0), (0, 1), (-1, 0)\}.
\end{split}
\end{equation}

Here, \( (x_0, y_0) \) is the spiral center, \( r_0 > 0 \) is the initial radius, and \( \alpha > 0 \) is the radial pitch. The direction vector \( \delta_{imod4} \) cycles incrementally in the four directions of up, right, down, and left.

In terms of implementation, the spiral path is encoded into several token streams, which are input into the Bi-WKV attention mechanism for iterative modeling. By applying the proposed Evolving Scanning mechanism to the RWKV structure, we ultimately bridge the gap between the Riemannian manifold properties of low-illumination edges and the temporal modeling capability of RWKV. This significantly improves edge continuity and structural fidelity under low-light conditions (the formula derivation process is provided in Appendix \textcolor{blue}{2}):

\begin{equation}
ES\text{-}RWKV(\cdot) = \cdot + \mathcal{F}_{\text{spatial}}(\cdot) + \mathcal{F}_{\text{channel}}(F_{\text{ESSM}}),
\end{equation}

where $\cdot$ denotes the input feature, \( \mathcal{F}_{\text{spatial}}(\cdot) \) represents the spatial dependency modeling implemented via spiral scanning and the ES-RWKV mechanism, and \( \mathcal{F}_{\text{channel}}(\cdot) \) denotes the channel correlation modeling based on Q-Shift and nonlinear mapping.

\subsection{Bilateral Spectrum Aligner Block}
\label{BSA}
Edge detail optimization can ensure feature focusing in low-light images, but its optimization degree and feature matching degree are particularly crucial. Specifically, excessive emphasis on details will lead to the prominence of noise in low-light regions and the generation of artifacts. Mismatch with brightness and color features will degrade visual perception. To address this issue, the spectral alignment technique proposed by us is embedded into DRWKV in the form of a Bilateral Spectrum Aligner Block (Bi-SAB).  

Structurally, Bi-SAB enables feature interaction between the brightness and color branches through CrossAttention. It utilizes Spectral 
Alignment Enhancer (SAE) to enhance image brightness and reduce noise, and extracts edge details with the Scharr operator. The complete structural block provides feasible solutions for issues related to brightness, color, details, and noise. The specific process is shown in Algorithm~\ref{alg:Bi-SAB}.

\begin{algorithm}[t]
\caption{Bi-SAB: Bilateral Spectrum Aligner Block}
\begin{algorithmic}[1]
\label{alg:Bi-SAB}
\item [\textbf{Input:}]
\parbox[t]{\textwidth}{
    Low-level Feature $F_{1}$: \textcolor{blue}{(B, $C_{i}$, $H_{i}$, $W_{i}$)}\\
    High-level Feature $F_{1}^{' }$: \textcolor{blue}{(B, $C_{i}$, $H_{i}$, $W_{i}$)}
}
\item [\textbf{Output:}] Bi-SAB Fusion Feature $F$: \textcolor{blue}{(B, $C_{i}$, $H_{i}$, $W_{i}$)}

\textcolor{red}{/*NOTE: Technical details are provided in the supplementary material.*/}

\STATE \textcolor{gray}{/*Compute the query (Q), key (K), and value (V) representations.*/}
\STATE $F_{Q}: \textcolor{blue}{(B, C_{i}, H_{i}, W_{i})} \gets \text{Conv}_{3 \times 3}(F_{1})$
\STATE $F_{K}: \textcolor{blue}{(B, C_{i}, H_{i}, W_{i})} \gets \text{Conv}_{3 \times 3}(F_{1}^{'})$
\STATE $F_{V}: \textcolor{blue}{(B, C_{1}, H_{i}, W_{i})} \gets \text{Conv}_{3 \times 3}(F_{1}^{'})$

\STATE \textcolor{gray}{/*Compute Cross Attention (CA) feature representations.*/}
\STATE $F_{attn}: \textcolor{blue}{(B, C_{1}, H_{i}, W_{i})} \gets CA(F_{Q},F_{K})$

\STATE \textcolor{gray}{/*Compute Feature Difference Adjustment (FDA) feature representations.*/}
\STATE $Att_{out}: \textcolor{blue}{(B, C_{1}, H_{i}, W_{i})} \gets F_{attn} \times F_{K}$
\STATE $F_{out}: \textcolor{blue}{(B, C_{1}, H_{i}, W_{i})} \gets \rho \cdot Att_{out} + F_{V}, \rho=0.2$
\STATE $FDA: \textcolor{blue}{(B, C_{1}, H_{i}, W_{i})} \gets F_{attn} \times F_{Q} + F_{out} $

\STATE \textcolor{gray}{/*Compute Scharr Edge Enhancement (SEE) feature representations.*/}
\STATE $F_{SEE}: \textcolor{blue}{(B, C_{i}, H_{i}, W_{i})} \gets SEE(F_{1})$

\STATE \textcolor{gray}{/*Compute Spectral Alignment Enhancer (SAE) feature representations.*/}
\STATE $F_{SAE}: \textcolor{blue}{(B, C_{1}, H_{i}, W_{i})} \gets SAE(LN(FDA))$

\STATE \textcolor{gray}{/*Compute Bi-SAB feature representations.*/}
\STATE $F: \textcolor{blue}{(B, C_{1}, H_{i}, W_{i})} \gets F_{SAE} + F_{SEE}$

\RETURN F
\end{algorithmic}
\end{algorithm}

\subsection{Loss Function}

\begin{table*}[htbp]
    \centering
    \fontsize{6.84pt}{10.2pt}\selectfont 
    \setlength{\tabcolsep}{2.4pt} 
    \renewcommand{\arraystretch}{1.2} 
    \caption{Experiment results on the LSRW-Huawei, LOLv2, and SDSD test benchmarks. The background color gradients from dark to light, with \cellcolor{red!80}{red} indicating the optimal result, \cellcolor{orange!80}{orange} denoting the sub-optimal result, and \cellcolor{yellow!80}{yellow} following in order.}
    \label{tab:performance_comparison}

    \begin{tabular*}{\textwidth}{l c c *{15}{c}}
        \toprule
        \multirow{2}{*}{\textbf{Method}} & 
        \multirow{2}{*}{\textbf{Params }} &  
        \multirow{2}{*}{\textbf{GFLOPs}} & 
        \multicolumn{3}{c}{\textbf{LSRW-Huawei}} & 
        \multicolumn{3}{c}{\textbf{LOLv2\_Real}} & 
        \multicolumn{3}{c}{\textbf{LOLv2\_Syn}} & 
        \multicolumn{3}{c}{\textbf{SDSD\_indoor}} & 
        \multicolumn{3}{c}{\textbf{SDSD\_outdoor}} \\
        & & & PSNR↑ & SSIM↑ & NIQE↓ & PSNR↑ & SSIM↑ & NIQE↓ & PSNR↑ & SSIM↑ & NIQE↓ & PSNR↑ & SSIM↑ & NIQE↓ & PSNR↑ & SSIM↑ & NIQE↓ \\
        \midrule
        RetinexNet~\cite{wei2018deep}       & 0.8 & 587.47 & 15.61 & 0.414 & 7.235 & 16.10 & 0.407 & 9.425 & 17.14 & 0.756 & 5.405 & 20.84 & 0.617 & 7.924 & 20.96 & 0.629 & 9.947 \\
        KinD~\cite{zhang2019kindling}       & 8.0 & 34.99  & 15.77 & 0.548 & 5.517 & 18.67 & 0.772 & 9.221 & 15.22 & 0.542 & 6.124 & 20.98 & 0.597 & 7.221 & 21.65 & 0.621 & 8.714 \\
        RRDNet~\cite{9102962}               & 0.1 & 2.10   & 14.66 & 0.541 & 7.725 & 15.21 & 0.514 & 9.912 & 16.67 & 0.667 & 4.894 & 21.07 & 0.601 & 6.921 & 20.04 & 0.647 & 8.997 \\
        Zero-DCE~\cite{guo2020zero}         & 0.7 & 5.21   & 14.86 & 0.559 & 4.215 & 14.12 & 0.512 & 8.652 & 14.93 & 0.531 & 5.507 & 21.23 & 0.752 & 6.941 & 20.39 & 0.691 & 8.014 \\
        SCI~\cite{ma2022toward}             & 0.09 & 0.06   & 14.78 & 0.526 & 3.667 & 17.30 & 0.541 & 8.077 & 14.96 & 0.721 & 4.899 & 20.43 & 0.714 & 6.574 & 21.33 & 0.701 & 7.778 \\
        \midrule
        LLFormer~\cite{wang2023ultra}       & 13.2 & 22.52  & 16.23 & 0.642 & 3.947 & 20.56 & 0.801 & 9.162 & 24.42 & 0.914 & 4.614 & 25.66 & 0.832 & 6.627 & 28.45 & 0.821 & 8.011 \\
        UHDFormer~\cite{wang2024correlation}& 0.3 & 48.37  & 21.07 & 0.604 & 4.621 & 21.59 & 0.804 & 7.230 & 22.60 & 0.903 & 5.476 & 28.13 & 0.875 & 7.021 & 22.75 & 0.732 & 6.657 \\
        Retinexformer~\cite{cai2023retinexformer}& 1.6 & 15.51 & \cellcolor{orange!20}22.24 & \cellcolor{yellow!20}0.701 & \cellcolor{orange!20}2.976 & 21.65 & 0.835 & \cellcolor{yellow!20}4.735 & \cellcolor{red!20}25.10 & 0.925 & \cellcolor{orange!20}3.971 & 28.96 & 0.879 & 5.441 & \cellcolor{red!20}28.96 & \cellcolor{red!20}0.896 & 7.201 \\
        IAT~\cite{cui2022you}               & 0.09 & 1.44   & 20.12 & 0.694 & 4.217 & 20.30 & 0.752 & 5.232 & 22.96 & 0.856 & 5.512 & 19.97 & 0.713 & 4.177 & 19.97 & 0.711 & 6.417 \\
        \midrule
        WalMaFa~\cite{Tan_2024_ACCV}        & 39.7 & 14.41  & 21.04 & 0.698 & 3.112 & \cellcolor{orange!20}22.49 & \cellcolor{red!20}0.851 & 4.365 & \cellcolor{red!20}25.10 & \cellcolor{orange!20}0.945 & 4.275 & \cellcolor{yellow!20}29.67 & \cellcolor{orange!20}0.915 & \cellcolor{orange!20}3.541 & \cellcolor{orange!20}28.94 & \cellcolor{orange!20}0.891 & \cellcolor{orange!20}3.955 \\ 
        MambaIR~\cite{guo2024mambair}       & 20.4 & 60.66  & \cellcolor{yellow!20}21.14 & \cellcolor{red!20}0.704 & \cellcolor{yellow!20}3.004 & 20.11 & 0.802 & 4.928 & 24.75 & 0.922 & 4.522 & 25.11 & 0.873 & \cellcolor{yellow!20}3.661 & 26.35 & 0.510 & 4.477 \\
        RetinexMamba~\cite{bai2024retinexmamba}& 4.6 & 34.75 & 20.88 & 0.629 & 3.104 & \cellcolor{yellow!20}22.34 & \cellcolor{yellow!20}0.826 & 4.771 & 24.71 & 0.932 & 4.503 & 28.21 & 0.893 & 3.976 & 26.22 & 0.866 & \cellcolor{yellow!20}4.011 \\
        MambaLLIE~\cite{weng2024mamballie}  & 4.4 & 20.85  & 20.64 & 0.627 & 3.047 & 22.14 & 0.821 & \cellcolor{orange!20}4.021 & \cellcolor{yellow!20}24.81 & \cellcolor{yellow!20}0.940 & \cellcolor{yellow!20}4.071 & \cellcolor{orange!20}29.74 & \cellcolor{yellow!20}0.902 & 4.417 & \cellcolor{yellow!20}28.92 & \cellcolor{yellow!20}0.869 & 4.417 \\
        \midrule
        \textbf{DRWKV (Ours)}               & 9.7 & 1.67   & \cellcolor{red!20}22.34 & \cellcolor{orange!20}0.703 & \cellcolor{red!20}2.944 & \cellcolor{red!20}24.12 & \cellcolor{orange!20}0.832 & \cellcolor{red!20}3.926 & \cellcolor{orange!20}25.02 & \cellcolor{red!20}0.947 & \cellcolor{red!20}3.941 & \cellcolor{red!20}30.26 & \cellcolor{red!20}0.922 & \cellcolor{red!20}3.441 & \cellcolor{red!20}28.96 & \cellcolor{orange!20}0.891 & \cellcolor{red!20}3.954 \\
        \bottomrule
    \end{tabular*}
\end{table*}

The newly defined MS$^{2}$-Loss focuses on five key attributes of the model: structure, edge, illumination, artifact, and weight.  

\noindent{\textbf{Decompositional consistency loss:}} Inspired by compressed sensing theory~\cite{doi:10.1126/science.1127647}, it applies the L1 norm to image reconstruction:  
\begin{equation}
\mathcal{L_{\text{recon}}}= \|I - \hat{I}\|_1,
\end{equation}  
where $\|\cdot\|_p$ denotes the $p$-norm. $\mathcal{L_{\text{recon}}}$ focuses on balancing pixel-level errors and structural information.  

\noindent{\textbf{Edge sparsity loss:}} The L1 loss is used to control the sparsity of edge contours and suppress false edges caused by noise:  
\begin{equation}
\mathcal{L_{\text{sparse}}} = \|E\|_1,
\end{equation}  
Herein, the edge map $E$ is non-zero only at real contours, while it should approach 0 at other positions, satisfying the sparsity prior condition.  

\noindent{\textbf{Illumination smoothness loss:}} In natural scenes, illumination $L$ exhibits characteristics of spatial smoothness and abrupt edge changes, requiring targeted handling:  
\begin{equation}
\mathcal{L_{\text{smooth}}} = |\nabla L| \cdot \exp(-\lambda |\nabla I|),
\end{equation}  
where $\nabla L$ is the illumination gradient, and $|\nabla I|$ represents the image region. A high $|\nabla I|$ indicates an edge region, where constraints on $L$ should be relaxed to emphasize illumination naturalness; a low $|\nabla I|$ indicates a flat region, where drastic changes in $L$ are penalized to avoid invalid fluctuations in illumination.  

\noindent{\textbf{Artifact suppression loss:}} A selection-regularized artifact suppression loss is proposed to minimize artifact energy and ensure spatial smoothness:  
\begin{equation}
\mathcal{L_{\text{artifact}}} = \|S\|_1 + \delta \cdot \text{TV}(S),
\end{equation}  
where $\delta$ is a weight parameter. When $\delta > 1$, $\mathcal{L}_{\text{noise}}$ is dominated by the TV regularization term, suppressing high-frequency oscillations such as halos and stripes; when $\delta < 1$, the loss is dominated by the L1 regularization term, controlling the magnitude of residuals.  

\noindent{{\textbf{Parameter regularization loss:}} Focusing on the optimization of learnable parameters $\alpha$, $\beta$, and $\gamma$, L2 regularization is equivalent to weight decay, forcing parameter values to approach 0, with the formula as follows:  
\begin{equation}
\mathcal{L_{\text{reg}}} = \alpha^2 + \beta^2 + \gamma^2.
\end{equation}  

\noindent{{\textbf{Total loss structure:}} The total loss structure is given by  
\begin{equation}
\begin{split}
\mathcal{L}_{\text{MS$^{2}$-Loss}} &= \lambda_1 \mathcal{L_{\text{recon}}} + \lambda_2 \mathcal{L_{\text{sparse}}} + \lambda_3 \mathcal{L_{\text{smooth}}} \\
&\quad + \lambda_4 \mathcal{L_{\text{artifact}}} + \lambda_5 \mathcal{L_{\text{reg}}},
\end{split}
\end{equation}  
where $\lambda_i$ are predefined weights.

\section{Experiments}
\subsection{Experimental Settings}
\noindent{\textbf{Implementation details:}}
We implement model training and inference on the NVIDIA A800-80GB platform. Specific parameter settings are as follows: the number of training epochs \( E \) is set to 500; the input image size \( S \) is 128×128; the block numbers of the DRWKV structure are set to \( N_1=4 \) and \( N_2=8 \) respectively, and the input channel number \( C \) is 16. The model optimizer uses Adam~\cite{kingma2014adam} with parameters set to \( \beta_1=0.9 \) and \( \beta_2=0.999 \); the initial learning rate is \( 2 \times 10^{-4} \), which is gradually reduced to \( 1 \times 10^{-6} \) via the cosine annealing strategy~\cite{loshchilov2016sgdr}.

\noindent{\textbf{Datasets:}}
\noindent\underline{\textit{(1) LSRW-Huawei}} is a subset of the LSRW~\cite{HAI2023103712} dataset consisting of images captured by Huawei smartphones, which includes 2450 pairs of training images and 30 pairs of test images. \noindent\underline{\textit{(2) LOLv2}}~\cite{9328179} is divided into two subsets: LOLv2 Real and LOLv2 Synthetic. The former is a dataset of real low-light scenes, containing 689 pairs of training images and 100 pairs of test images, covering various environments such as campuses and clubs. The latter is a synthetic dataset, consisting of 900 pairs of training images and 100 pairs of test images. \noindent\underline{\textit{(3) SDSD}}~\cite{9710730} dataset has two subsets: indoor and outdoor. The indoor subset includes 62 pairs of low-light/normal-light videos for training and 6 pairs of low-light/normal-light videos for testing. The outdoor subset contains 116 pairs of low-light/normal-light videos for training and 10 pairs of low-light/normal-light videos for testing.

\noindent{\textbf{{Metrics:}}
We use traditional quality (distortion) metrics PSNR and SSIM~\cite{1284395} to evaluate the results of LLIE, aiming to measure the pixel-level similarity and distortion between the enhanced images and reference images. Additionally, NIQE~\cite{6353522} is employed to assess the no-reference perceptual quality of the enhanced images.

\subsection{Comparative Experiments}
\noindent\textbf{Model Comparison Results:}
\begin{figure*}[t]
    \centering
    \includegraphics[width=\linewidth]{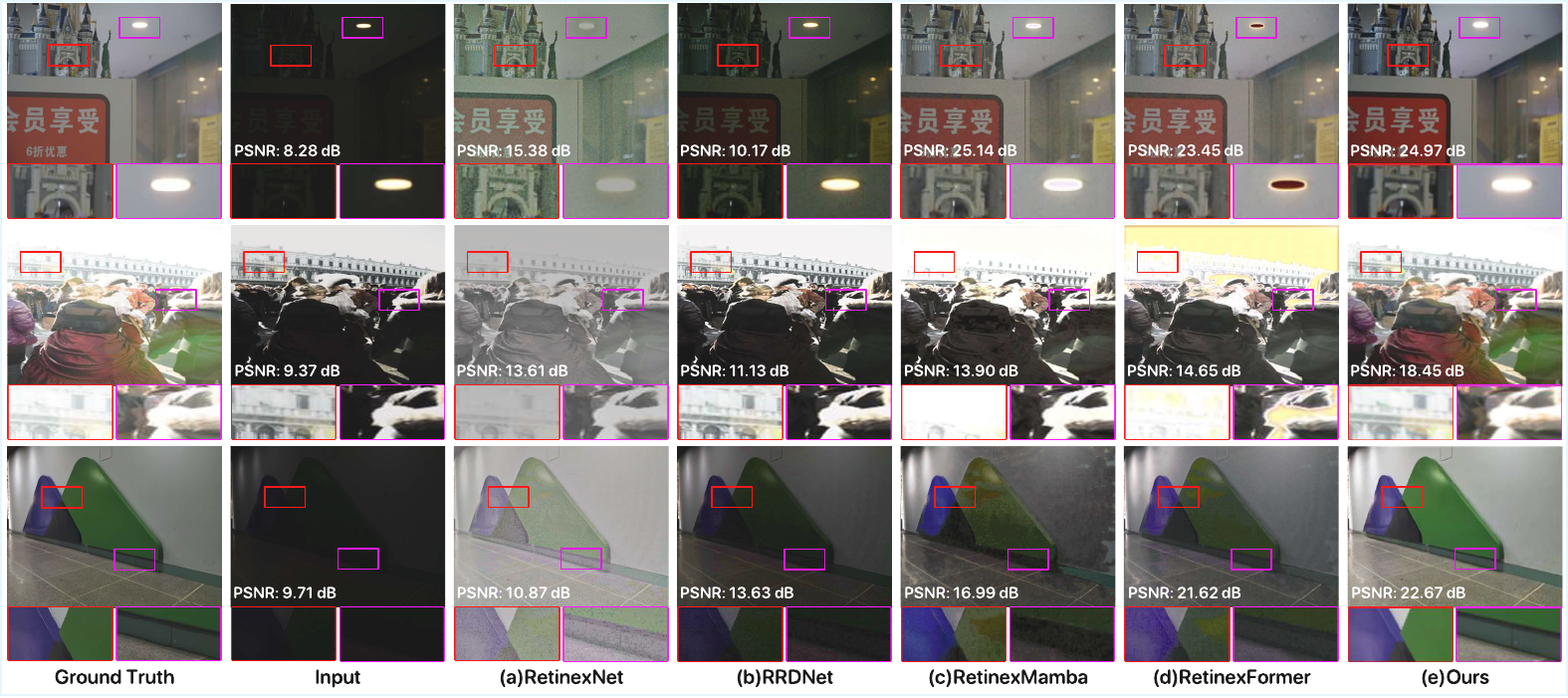}
    \caption{Visualization results of RetinexNet, RRDNet, RetinexMamba, RetinexFormer, and DRWKV on the LOLv2-Real, LOLv2-Synthetic, and LSRW-Huawei datasets respectively (zoom in to display details for each model).}
    \label{fig:block_comparison}
\end{figure*}
Based on the quantitative results across five test benchmarks, the DRWKV model achieves performance improvements over multiple baseline methods with extremely low GFLOPS, as shown in Table~\ref{tab:performance_comparison}. Specifically, DRWKV exhibits superior performance in terms of naturalness on \textbf{LSRW–Huawei}, characterized by low baseline values of NIQE and a high optimization rate (achieves a 1.1\% optimization compared to the second-place method). Additionally, on \textbf{LOLv2}, DRWKV shows differences in optimization strengths between the Real and Synthetic test benchmarks: it achieves the optimal model performance with an average improvement of 5.00 dB in PSNR on Real and 0.152 in SSIM on Synthetic, balancing the enhancement of image quality and the preservation of structural information. In low-light scenario videos, by independently enhancing low-light frames in \textbf{SDSD}, the model demonstrates average PSNR differences of up to 9.42 dB/8 dB in indoor and outdoor scenes, ensuring high ``restoration accuracy" of image details. For visualization results, refer to Figure~\ref{fig:block_comparison}; the generated results of our DRWKV model exhibit continuous object edges and high-fidelity details, which are highly consistent with the subject-focused edges.
\begin{table}[htbp]
    \centering
    \fontsize{8.2pt}{10.2pt}\selectfont 
    \setlength{\tabcolsep}{3.2pt}     
    \renewcommand{\arraystretch}{1.2} 
    \caption{Comparison of different Block combinations on the LOLv2 Real benchmark dataset.}
    \label{tab:block_comparison}
    \begin{tabular}{lccccc}
        \toprule
        \textbf{Config} & \textbf{Block1} & \textbf{Block2} & \textbf{PSNR↑} & \textbf{SSIM↑} & \textbf{NIQE↓} \\
        \midrule
        $(\mathrm{I})$ & Robust Retinex & Robust Retinex & 22.57 & 0.741 & 4.124 \\
        \textbf{Ours} & Robust Retinex & GER & \cellcolor{red!20}24.12 & \cellcolor{red!20}0.832 & \cellcolor{red!20}3.926 \\
        \bottomrule
    \end{tabular}
\end{table}

\noindent\textbf{Retinex Configuration Results:} 
This experiment focuses on the contradiction of edge distortion under low light. In terms of network structure, replaceable structures Block1 and Block2 are preset to assist in configuring Retinex, enabling DRWKV to achieve global optimum. Since the input end of the GER theory needs to receive the edge feature \( E \) (a parameter from the output end of DRWKV), Block1 is constrained by structural limitations. To minimize disturbance to the network architecture, one-sided fixation is ultimately adopted. Deploying the GER theory into Block2 results in remarkable improvement rates of 6.9\%/12.3\%/4.8\% for PSNR, SSIM, and NIQE, respectively, as shown in Table~\ref{tab:block_comparison}. This strongly validates the core idea of gradient-guided global-edge coupling in the GER theory, demonstrating its strong practicality in addressing edge issues of low-light images and further confirming \textit{Hypothesis(1)}.

\subsection{Ablation Studies}

\noindent\textbf{Loss Performance Results:} 
A targeted ablation study was performed on the proposed loss function combinations. In Figure~\ref{fig:loss}\hyperlink{fig:intro-a}{(a)}, the absence of the decompositional consistency loss $\mathcal{L_{\text{recon}}}$ results in an overall darker reconstructed image with incomplete edge information, indirectly validating its role in structural and luminance reconstruction. In Figure~\ref{fig:loss}\hyperlink{fig:intro-b}{(b)}, removing the edge sparsity loss $\mathcal{L_{\text{sparse}}}$ leads to false edges generated by Gaussian noise contaminating flat regions, highlighting its contribution to preserving and enhancing low-light edges. Figures~\ref{fig:loss}\hyperlink{fig:intro-c}{(c)} and~\ref{fig:loss}\hyperlink{fig:intro-d}{(d)} illustrate uneven illumination and over-denoising effects caused by the removal of the illumination smoothness loss $\mathcal{L_{\text{smooth}}}$ and artifact suppression loss $\mathcal{L_{\text{artifact}}}$, respectively, confirming their balancing roles. Figure~\ref{fig:loss}\hyperlink{fig:intro-e}{(e)} demonstrates model instability with negative image artifacts when the parameter regularization loss $\mathcal{L_{\text{reg}}}$ is removed, indicating that effective regularization can improve the model's adaptability to unseen data. Additionally, the loss parameters for this model are specified as follows: \( \lambda_1=1 \), \( \lambda_2=0.01 \), \( \lambda_3=0.1 \), \( \lambda_4=0.05 \), and \( \lambda_5=1e-4 \).

\begin{figure*}[t]
    \centering
    \includegraphics[width=\linewidth]{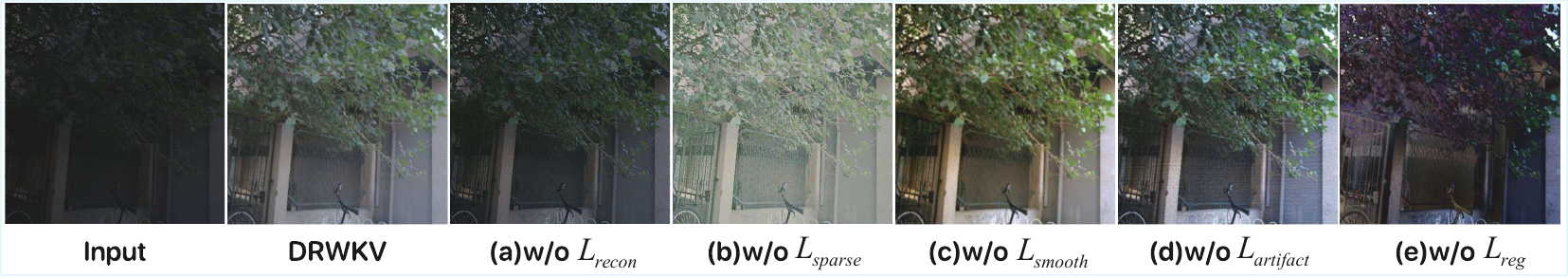}
    \caption{An ablation study is conducted on the proposed decompositional consistency loss \( L_{recon} \), edge sparsity loss \( L_{sparse} \), illumination smoothness loss \( L_{smooth} \), artifact suppression loss \( L_{artifact} \), and parameter regularization loss \( L_{reg} \).}
    \hypertarget{fig:intro-a}{}
    \hypertarget{fig:intro-b}{}
    \hypertarget{fig:intro-c}{}
    \hypertarget{fig:intro-d}{}
    \hypertarget{fig:intro-e}{}
    \label{fig:loss}
\end{figure*}

\noindent\textbf{Component Ablation Results:} 
This experiment aims to verify the effectiveness of the Evolving Scanning mechanism proposed in \textit{Hypothesis(2)} for the edges of low-light images. Structurally, a progressive component integration strategy is adopted, with LOLv2 Real as the test benchmark and the basic structure as the test framework, to construct a five-level progressive validation system. The component configurations, ablation variables, and validation results at each level are shown in Table~\ref{tab:component_analysis}. In terms of the basic structure, we retain the ViT-style block stacking structure of VRWKV and introduce the SSIM+L1 loss, enabling it to initially possess the capability to handle low-light issues. However, the model exhibits nearly ineffective feature extraction in illumination-degraded scenarios, characterized by extremely low SSIM (0.415) and PSNR (12.57 dB) values.  

Validation (1): After embedding the Evolving Scanning module (ES-RWKV), the performance of SSIM and PSNR increased by 42.4\% and 29.4\%, respectively. This explicitly verifies that the proposed spiral scanning is compatible with the edge spatial topology, capable of ensuring the continuity of edges of low-light objects.  

Validation (2): By modifying the network from the ViT-style block stacking scheme to a UNet-like scheme, the model performance optimization focuses on detail accuracy. At the cost of a certain increase in parameter count and computational load, the PSNR index gains 3.16 dB.  

Validation (3): Adding a light preprocessing(LP) stage leads to a slight increase in image quality evaluation metrics, demonstrating that this stage contributes to improving pixel errors and structures.  

Validation (4): Embedding the Bi-SAB module and adopting high-performance multi-module cascading significantly optimize the model structure. The parameter count and computational load are drastically reduced by 49.3\% and 43.8\%, respectively, while SSIM and PSNR achieve beneficial gains of 0.028 and 0.8 dB, endowing the model with basic lightweight characteristics.  

Validation (5): Replacing the SSIM+L1 loss with the proposed MS$^{2}$-Loss specifically optimizes the training process, with SSIM and PSNR optimized to 0.832 and 24.12 dB, respectively.  

At this point, the hypotheses proposed in the Methodology have been fully validated. The paper constructs a complete research closed-loop from four dimensions: theory, technology, experiment, and application.

\begin{table}[htbp]
    \centering
    \fontsize{7pt}{10.2pt}\selectfont 
    \setlength{\tabcolsep}{2.3pt}  
    \renewcommand{\arraystretch}{1.2}
    \caption{Ablation study on the LOLv2 Real benchmark dataset. If the RWKV blocks and training losses are not selected, Bi-RWKV and SSIM+L1 loss will be used by default. \ding{52}\rotatebox[origin=c]{-9.2}{\kern-0.7em\ding{55}} denotes the use of a ViT-style block stacking structure, whereas \ding{52} indicates the adoption of a UNet-like architecture.}
    \label{tab:component_analysis}
    \begin{tabular}{cccccccc}
        \toprule
        \textbf{ES-RWKV} & \textbf{LP} & \textbf{Bi-SAB} & \textbf{MS$^{2}$-Loss} & \textbf{SSIM↑} & \textbf{PSNR↑} & \textbf{Params.(M)} & \textbf{GFLOPs} \\
        \midrule
         & & & & 0.415 & 12.57 & 10.51 & 0.02 \\
        \ding{52}\rotatebox[origin=c]{-9.2}{\kern-0.7em\ding{55}} & & & & 0.591 & 16.27 & 14.57 & 2.65 \\     
        \ding{52} & & & & 0.607 & 19.43 & 16.36 & 2.95 \\
        \ding{52} & \ding{52} & & & 0.748 & 22.67 & 16.36 & 2.97 \\
        \ding{52} & \ding{52} & \ding{52} & & 0.776 & 23.47 & 8.29 & 1.67 \\
        \ding{52} & \ding{52} & \ding{52} & \ding{52} & 0.832 & 24.12 & 8.28 & 1.67 \\
        \bottomrule
    \end{tabular}
\end{table}

\subsection{Extended Validation}

To evaluate generalization, we applied DRWKV to low-light multi-object tracking on the UAVDark135 benchmark~\cite{li2021allday}, comprising 135 sequences across diverse scenarios at 30 FPS and 1920×1080 resolution.

Mainstream tracking algorithms were tested with and without DRWKV, using MOTA, IDF1, and HOTA as metrics. DRWKV consistently improved performance, notably in identity preservation. ByteTrack saw gains of 5.189\%/3.076\%/2.705\%, while BoT-SORT and others improved by 1.5\%–7\%.

In contrast, Deep OC-SORT showed mixed results, with improved MOTA but reduced IDF1, suggesting structural sensitivity to enhanced features. Implementation details are provided in the Appendix.

\begin{table}[htbp]
    \centering
    \fontsize{6.7pt}{10.2pt}\selectfont 
    \setlength{\tabcolsep}{1.4pt}     
    \renewcommand{\arraystretch}{1.2} 
    \caption{Extended validation results of DRWKV on low-light object tracking performance. MOTA~\cite{bernardin2008evaluating}, IDF1~\cite{ristani2016performance} and HOTA~\cite{luiten2021hota} are common performance evaluation metrics for object tracking.}
    \label{tab:tracking_performance}
    \begin{tabular}{lcccccc}
        \toprule
        \multirow{2}{*}{\textbf{Methods}} & \multicolumn{3}{c}{\textbf{w/o LLIE Module}} & \multicolumn{3}{c}{\textbf{DRWKV (Ours)}} \\
        & MOTA(\%) & IDF1(\%) & HOTA(\%) & MOTA(\%) & IDF1(\%) & HOTA(\%) \\
        \midrule
        ByteTrack~\cite{zhang2022bytetrack} & 23.434 & 62.776 & 50.117 & \cellcolor{orange!20}28.623 & \cellcolor{yellow!20}65.852 & \cellcolor{orange!20}52.822 \\
        BoT-SORT~\cite{aharon2022bot} & 18.242 & 62.916 & 49.556 & 25.496 & 64.049 & 51.435 \\
        Hybrid-SORT~\cite{yang2024hybrid} & 20.684 & 62.054 & 49.196 & \cellcolor{red!20}29.355 & \cellcolor{red!20}66.365 & \cellcolor{red!20}52.905 \\
        OC-SORT~\cite{Cao_2023_CVPR} & 21.062 & 62.967 & 49.469 & \cellcolor{yellow!20}28.102 & \cellcolor{orange!20}66.000 & \cellcolor{yellow!20}52.501 \\
        Deep OC-SORT~\cite{maggiolino2023deep} & 19.514 & 63.205 & 49.857 & 24.929 & 58.016 & 49.124 \\
        StrongSORT~\cite{10032656} & 20.923 & 53.583 & 45.553 & 25.148 & 64.677 & 51.842 \\
        \bottomrule
    \end{tabular}
\end{table}


\section{Conclusion}
In this study, we introduced the DRWKV model to address the challenge of preserving edge details in Low-Light Image Enhancement (LLIE). By integrating the Global Edge Retinex (GER) theory, the model decouples illumination and edge structures, ensuring high-fidelity edge continuity even in extreme low-light conditions. The Evolving WKV Attention mechanism enhances spatial edge continuity, while the Bilateral Spectrum Aligner Block (Bi-SAB) and MS$^{2}$-Loss optimize luminance and chrominance alignment, reducing artifacts and improving visual quality. Extensive experiments on benchmark datasets show that DRWKV outperforms existing methods in PSNR, SSIM, and NIQE, with lower computational cost. Additionally, it proves effective in tasks like low-light multi-object tracking, enhancing performance, especially in identity preservation. Future work will focus on adapting DRWKV for mobile platforms and further improving its robustness across various low-light scenarios, bridging the gap between academic research and real-world deployment.

{\small
\bibliographystyle{ieee_fullname}
\bibliography{egbib}

@String(CVPR= {IEEE Conf. Comput. Vis. Pattern Recog.})

@String(ICCV= {Int. Conf. Comput. Vis.})

@String(ICME = {Int. Conf. Multimedia and Expo})

@String(ICIP = {IEEE Int. Conf. Image Process.})

@String(ACCV  = {ACCV})

@String(AAAI = {AAAI})

@String(CVPR  = {CVPR})

@String(ICCV  = {ICCV})

@String(ICME  =	{ICME})

@String(ICIP  = {ICIP})

@article{xu2025upt,
  title={UPT-Flow: Multi-scale transformer-guided normalizing flow for low-light image enhancement},
  author={Xu, Lintao and Hu, Changhui and Hu, Yin and Jing, Xiaoyuan and Cai, Ziyun and Lu, Xiaobo},
  journal={Pattern Recognition},
  volume={158},
  pages={111076},
  year={2025},
  publisher={Elsevier}
}

@article{wang2024low,
  title={Low-light wheat image enhancement using an explicit inter-channel sparse transformer},
  author={Wang, Yu and Wang, Fei and Li, Kun and Feng, Xuping and Hou, Wenhui and Liu, Lu and Chen, Liqing and He, Yong and Wang, Yuwei},
  journal={Computers and Electronics in Agriculture},
  volume={224},
  pages={109169},
  year={2024},
  publisher={Elsevier}
}

@inproceedings{shi2024zero,
  title={Zero-ig: zero-shot illumination-guided joint denoising and adaptive enhancement for low-light images},
  author={Shi, Yiqi and Liu, Duo and Zhang, Liguo and Tian, Ye and Xia, Xuezhi and Fu, Xiaojing},
  booktitle={Proceedings of the IEEE/CVF conference on computer vision and pattern recognition},
  pages={3015--3024},
  year={2024}
}

@inproceedings{wang2024multimodal,
  title={Multimodal low-light image enhancement with depth information},
  author={Wang, Zhen and Li, Dongyuan and Li, Guang and Zhang, Ziqing and Jiang, Renhe},
  booktitle={Proceedings of the 32nd ACM International Conference on Multimedia},
  pages={4976--4985},
  year={2024}
}

@inproceedings{zou2024wave,
  title={Wave-mamba: Wavelet state space model for ultra-high-definition low-light image enhancement},
  author={Zou, Wenbin and Gao, Hongxia and Yang, Weipeng and Liu, Tongtong},
  booktitle={Proceedings of the 32nd ACM International Conference on Multimedia},
  pages={1534--1543},
  year={2024}
}

@article{shang2024holistic,
  title={Holistic dynamic frequency transformer for image fusion and exposure correction},
  author={Shang, Xiaoke and Li, Gehui and Jiang, Zhiying and Zhang, Shaomin and Ding, Nai and Liu, Jinyuan},
  journal={Information Fusion},
  volume={102},
  pages={102073},
  year={2024},
  publisher={Elsevier}
}

@article{qu2024double,
  title={Double domain guided real-time low-light image enhancement for ultra-high-definition transportation surveillance},
  author={Qu, Jingxiang and Liu, Ryan Wen and Gao, Yuan and Guo, Yu and Zhu, Fenghua and Wang, Fei-Yue},
  journal={IEEE Transactions on Intelligent Transportation Systems},
  volume={25},
  number={8},
  pages={9550--9562},
  year={2024},
  publisher={IEEE}
}

@inproceedings{zhang2024cnn,
  title={A Cnn-Transformer Network Based Snr Guided High Frequency Reconstruction for Low Light Image Enhancement},
  author={Zhang, Jin and Jin, Haiyan and Su, Haonan and Zhang, Yuanlin and Xiao, Zhaolin and Wang, Bin},
  booktitle={2024 IEEE International Conference on Image Processing (ICIP)},
  pages={1649--1655},
  year={2024},
  organization={IEEE}
}

@inproceedings{nguyen2024diffusion,
  title={Diffusion in the dark: A diffusion model for low-light text recognition},
  author={Nguyen, Cindy M and Chan, Eric R and Bergman, Alexander W and Wetzstein, Gordon},
  booktitle={Proceedings of the IEEE/CVF Winter Conference on Applications of Computer Vision},
  pages={4146--4157},
  year={2024}
}

@article{guo2016lime,
  title={LIME: Low-light image enhancement via illumination map estimation},
  author={Guo, Xiaojie and Li, Yu and Ling, Haibin},
  journal={IEEE Transactions on image processing},
  volume={26},
  number={2},
  pages={982--993},
  year={2016},
  publisher={IEEE}
}

@article{chen2024highly,
  title={Highly robust thermal infrared and visible image registration with canny and phase congruence detection},
  author={Chen, Mengting and Yi, Shi and Wu, Lang and Yin, Hongli and Chen, Ling},
  journal={Optics and Lasers in Engineering},
  volume={183},
  pages={108526},
  year={2024},
  publisher={Elsevier}
}

@article{lu2022low,
  title={Low-light image enhancement via gradient prior-aided network},
  author={Lu, Yuxu and Gao, Yuan and Guo, Yongqi and Xu, Wenyu and Hu, Xianjun},
  journal={IEEE Access},
  volume={10},
  pages={92583--92596},
  year={2022},
  publisher={IEEE}
}

@article{lim2020dslr,
  title={DSLR: Deep stacked Laplacian restorer for low-light image enhancement},
  author={Lim, Seokjae and Kim, Wonjun},
  journal={IEEE Transactions on Multimedia},
  volume={23},
  pages={4272--4284},
  year={2020},
  publisher={IEEE}
}

@article{wang2022eanet,
  title={EANet: Iterative edge attention network for medical image segmentation},
  author={Wang, Kun and Zhang, Xiaohong and Zhang, Xiangbo and Lu, Yuting and Huang, Sheng and Yang, Dan},
  journal={Pattern Recognition},
  volume={127},
  pages={108636},
  year={2022},
  publisher={Elsevier}
}

@article{zhao2024dyedgegat,
  title={Dyedgegat: Dynamic edge via graph attention for early fault detection in iiot systems},
  author={Zhao, Mengjie and Fink, Olga},
  journal={IEEE Internet of Things Journal},
  year={2024},
  publisher={IEEE}
}

@inproceedings{li2023edge,
  title={Edge-aware regional message passing controller for image forgery localization},
  author={Li, Dong and Zhu, Jiaying and Wang, Menglu and Liu, Jiawei and Fu, Xueyang and Zha, Zheng-Jun},
  booktitle={Proceedings of the IEEE/CVF Conference on Computer Vision and Pattern Recognition},
  pages={8222--8232},
  year={2023}
}

@article{tang2025spiralmamba,
  title={SpiralMamba: Spatial-Spectral Complementary Mamba with Spatial Spiral Scan for Hyperspectral Image Classification},
  author={Tang, Xu and Yao, Yuexi and Ma, Jingjing and Zhang, Xiangrong and Yang, Yuqun and Wang, Bo and Jiao, Licheng},
  journal={IEEE Transactions on Geoscience and Remote Sensing},
  year={2025},
  publisher={IEEE}
}

@article{lore2017llnet,
  title={LLNet: A deep autoencoder approach to natural low-light image enhancement},
  author={Lore, Kin Gwn and Akintayo, Adedotun and Sarkar, Soumik},
  journal={Pattern Recognition},
  volume={61},
  pages={650--662},
  year={2017},
  publisher={Elsevier}
}

@inproceedings{pu2022edter,
  title={Edter: Edge detection with transformer},
  author={Pu, Mengyang and Huang, Yaping and Liu, Yuming and Guan, Qingji and Ling, Haibin},
  booktitle={Proceedings of the IEEE/CVF conference on computer vision and pattern recognition},
  pages={1402--1412},
  year={2022}
}

@article{chen2024retinex,
  title={Retinex-RAWMamba: Bridging Demosaicing and Denoising for Low-Light RAW Image Enhancement},
  author={Chen, Xianmin and Huang, Peiliang and Feng, Xiaoxu and Zhang, Dingwen and Han, Longfei and Han, Junwei},
  journal={arXiv preprint arXiv:2409.07040},
  year={2024}
}

@article{guan2021dual,
  title={A dual-tree complex wavelet transform-based model for low-illumination image enhancement},
  author={GUAN, Yurong and Aamir, Muhammad and Rahman, Ziaur and Dayo, Zaheer Ahmed and Abro, Waheed Ahmed and Ishfaq, Muhammad and HU, Zhihua},
  journal={Wuhan University Journal of Natural Sciences},
  volume={26},
  number={05},
  pages={405--414},
  year={2021}
}

@article{qian2021improved,
  title={Improved infrared image edge detection algorithm based on DexiNed},
  author={Qian, HE and Boyun, LIU},
  journal={Infrared Technology},
  volume={43},
  number={9},
  pages={876--884},
  year={2021}
}

@inproceedings{liu2024optical,
  title={Optical Flow-Guided 6DoF Object Pose Tracking with an Event Camera},
  author={Liu, Zibin and Guan, Banglei and Shang, Yang and Liang, Shunkun and Yu, Zhenbao and Yu, Qifeng},
  booktitle={Proceedings of the 32nd ACM International Conference on Multimedia},
  pages={6501--6509},
  year={2024}
}

@inproceedings{bai2024retinexmamba,
  title={Retinexmamba: Retinex-based mamba for low-light image enhancement},
  author={Bai, Jiesong and Yin, Yuhao and He, Qiyuan and Li, Yuanxian and Zhang, Xiaofeng},
  booktitle={International Conference on Neural Information Processing},
  pages={427--442},
  year={2024},
  organization={Springer}
}

@article{duan2024vision,
  title={Vision-rwkv: Efficient and scalable visual perception with rwkv-like architectures},
  author={Duan, Yuchen and Wang, Weiyun and Chen, Zhe and Zhu, Xizhou and Lu, Lewei and Lu, Tong and Qiao, Yu and Li, Hongsheng and Dai, Jifeng and Wang, Wenhai},
  journal={arXiv preprint arXiv:2403.02308},
  year={2024}
}

@ARTICLE{8304597,
  author={Li, Mading and Liu, Jiaying and Yang, Wenhan and Sun, Xiaoyan and Guo, Zongming},
  journal={IEEE Transactions on Image Processing}, 
  title={Structure-Revealing Low-Light Image Enhancement Via Robust Retinex Model}, 
  year={2018},
  volume={27},
  number={6},
  pages={2828-2841},
  doi={10.1109/TIP.2018.2810539}}

@article{kingma2014adam,
  title={Adam: A method for stochastic optimization},
  author={Kingma, Diederik P and Ba, Jimmy},
  journal={arXiv preprint arXiv:1412.6980},
  year={2014}
}

@article{loshchilov2016sgdr,
  title={Sgdr: Stochastic gradient descent with warm restarts},
  author={Loshchilov, Ilya and Hutter, Frank},
  journal={arXiv preprint arXiv:1608.03983},
  year={2016}
}

@ARTICLE{1284395,
  author={Zhou Wang and Bovik, A.C. and Sheikh, H.R. and Simoncelli, E.P.},
  journal={IEEE Transactions on Image Processing}, 
  title={Image quality assessment: from error visibility to structural similarity}, 
  year={2004},
  volume={13},
  number={4},
  pages={600-612},
  doi={10.1109/TIP.2003.819861}}

@ARTICLE{6353522,
  author={Mittal, Anish and Soundararajan, Rajiv and Bovik, Alan C.},
  journal={IEEE Signal Processing Letters}, 
  title={Making a “Completely Blind” Image Quality Analyzer}, 
  year={2013},
  volume={20},
  number={3},
  pages={209-212},
  doi={10.1109/LSP.2012.2227726}}

@article{bernardin2008evaluating,
  title={Evaluating multiple object tracking performance: the clear mot metrics},
  author={Bernardin, Keni and Stiefelhagen, Rainer},
  journal={EURASIP Journal on Image and Video Processing},
  volume={2008},
  number={1},
  pages={246309},
  year={2008},
  publisher={Springer}
}

@inproceedings{ristani2016performance,
  title={Performance measures and a data set for multi-target, multi-camera tracking},
  author={Ristani, Ergys and Solera, Francesco and Zou, Roger and Cucchiara, Rita and Tomasi, Carlo},
  booktitle={European conference on computer vision},
  pages={17--35},
  year={2016},
  organization={Springer}
}

@article{luiten2021hota,
  title={Hota: A higher order metric for evaluating multi-object tracking},
  author={Luiten, Jonathon and Osep, Aljosa and Dendorfer, Patrick and Torr, Philip and Geiger, Andreas and Leal-Taix{\'e}, Laura and Leibe, Bastian},
  journal={International journal of computer vision},
  volume={129},
  number={2},
  pages={548--578},
  year={2021},
  publisher={Springer}
}

@article{wei2018deep,
  title={Deep retinex decomposition for low-light enhancement},
  author={Wei, Chen and Wang, Wenjing and Yang, Wenhan and Liu, Jiaying},
  journal={arXiv preprint arXiv:1808.04560},
  year={2018}
}

@inproceedings{zhang2019kindling,
  title={Kindling the darkness: A practical low-light image enhancer},
  author={Zhang, Yonghua and Zhang, Jiawan and Guo, Xiaojie},
  booktitle={Proceedings of the 27th ACM international conference on multimedia},
  pages={1632--1640},
  year={2019}
}

@inproceedings{guo2020zero,
  title={Zero-reference deep curve estimation for low-light image enhancement},
  author={Guo, Chunle and Li, Chongyi and Guo, Jichang and Loy, Chen Change and Hou, Junhui and Kwong, Sam and Cong, Runmin},
  booktitle={Proceedings of the IEEE/CVF conference on computer vision and pattern recognition},
  pages={1780--1789},
  year={2020}
}

@inproceedings{ma2022toward,
  title={Toward fast, flexible, and robust low-light image enhancement},
  author={Ma, Long and Ma, Tengyu and Liu, Risheng and Fan, Xin and Luo, Zhongxuan},
  booktitle={Proceedings of the IEEE/CVF conference on computer vision and pattern recognition},
  pages={5637--5646},
  year={2022}
}

@inproceedings{wang2023ultra,
  title={Ultra-high-definition low-light image enhancement: A benchmark and transformer-based method},
  author={Wang, Tao and Zhang, Kaihao and Shen, Tianrun and Luo, Wenhan and Stenger, Bjorn and Lu, Tong},
  booktitle={Proceedings of the AAAI conference on artificial intelligence},
  volume={37},
  number={3},
  pages={2654--2662},
  year={2023}
}

@inproceedings{wang2024correlation,
  title={Correlation matching transformation transformers for uhd image restoration},
  author={Wang, Cong and Pan, Jinshan and Wang, Wei and Fu, Gang and Liang, Siyuan and Wang, Mengzhu and Wu, Xiao-Ming and Liu, Jun},
  booktitle={Proceedings of the AAAI Conference on Artificial Intelligence},
  volume={38},
  number={6},
  pages={5336--5344},
  year={2024}
}

@inproceedings{cai2023retinexformer,
  title={Retinexformer: One-stage retinex-based transformer for low-light image enhancement},
  author={Cai, Yuanhao and Bian, Hao and Lin, Jing and Wang, Haoqian and Timofte, Radu and Zhang, Yulun},
  booktitle={Proceedings of the IEEE/CVF international conference on computer vision},
  pages={12504--12513},
  year={2023}
}

@article{cui2022you,
  title={You only need 90k parameters to adapt light: a light weight transformer for image enhancement and exposure correction},
  author={Cui, Ziteng and Li, Kunchang and Gu, Lin and Su, Shenghan and Gao, Peng and Jiang, Zhengkai and Qiao, Yu and Harada, Tatsuya},
  journal={arXiv preprint arXiv:2205.14871},
  year={2022}
}

@inproceedings{guo2024mambair,
  title={Mambair: A simple baseline for image restoration with state-space model},
  author={Guo, Hang and Li, Jinmin and Dai, Tao and Ouyang, Zhihao and Ren, Xudong and Xia, Shu-Tao},
  booktitle={European conference on computer vision},
  pages={222--241},
  year={2024},
  organization={Springer}
}

@InProceedings{Tan_2024_ACCV,
    author    = {Tan, Junhao and Pei, Songwen and Qin, Wei and Fu, Bo and Li, Ximing and Huang, Libo},
    title     = {Wavelet-based Mamba with Fourier Adjustment for Low-light Image Enhancement},
    booktitle = {Proceedings of the Asian Conference on Computer Vision (ACCV)},
    month     = {December},
    year      = {2024},
    pages     = {3449-3464}
}

@article{weng2024mamballie,
  title={Mamballie: Implicit retinex-aware low light enhancement with global-then-local state space},
  author={Weng, Jiangwei and Yan, Zhiqiang and Tai, Ying and Qian, Jianjun and Yang, Jian and Li, Jun},
  journal={Advances in Neural Information Processing Systems},
  volume={37},
  pages={27440--27462},
  year={2024}
}

@inproceedings{zhang2022bytetrack,
  title={Bytetrack: Multi-object tracking by associating every detection box},
  author={Zhang, Yifu and Sun, Peize and Jiang, Yi and Yu, Dongdong and Weng, Fucheng and Yuan, Zehuan and Luo, Ping and Liu, Wenyu and Wang, Xinggang},
  booktitle={European conference on computer vision},
  pages={1--21},
  year={2022},
  organization={Springer}
}

@article{aharon2022bot,
  title={BoT-SORT: Robust associations multi-pedestrian tracking},
  author={Aharon, Nir and Orfaig, Roy and Bobrovsky, Ben-Zion},
  journal={arXiv preprint arXiv:2206.14651},
  year={2022}
}

@inproceedings{yang2024hybrid,
  title={Hybrid-sort: Weak cues matter for online multi-object tracking},
  author={Yang, Mingzhan and Han, Guangxin and Yan, Bin and Zhang, Wenhua and Qi, Jinqing and Lu, Huchuan and Wang, Dong},
  booktitle={Proceedings of the AAAI conference on artificial intelligence},
  volume={38},
  number={7},
  pages={6504--6512},
  year={2024}
}

@ARTICLE{10032656,
  author={Du, Yunhao and Zhao, Zhicheng and Song, Yang and Zhao, Yanyun and Su, Fei and Gong, Tao and Meng, Hongying},
  journal={IEEE Transactions on Multimedia}, 
  title={StrongSORT: Make DeepSORT Great Again}, 
  year={2023},
  volume={25},
  number={},
  pages={8725-8737},
  doi={10.1109/TMM.2023.3240881}}

@InProceedings{Cao_2023_CVPR,
    author    = {Cao, Jinkun and Pang, Jiangmiao and Weng, Xinshuo and Khirodkar, Rawal and Kitani, Kris},
    title     = {Observation-Centric SORT: Rethinking SORT for Robust Multi-Object Tracking},
    booktitle = {Proceedings of the IEEE/CVF Conference on Computer Vision and Pattern Recognition (CVPR)},
    month     = {June},
    year      = {2023},
    pages     = {9686-9696}
}

@inproceedings{maggiolino2023deep,
  title={Deep oc-sort: Multi-pedestrian tracking by adaptive re-identification},
  author={Maggiolino, Gerard and Ahmad, Adnan and Cao, Jinkun and Kitani, Kris},
  booktitle={2023 IEEE International conference on image processing (ICIP)},
  pages={3025--3029},
  year={2023},
  organization={IEEE}
}

@INPROCEEDINGS{9102962,
  author={Zhu, Anqi and Zhang, Lin and Shen, Ying and Ma, Yong and Zhao, Shengjie and Zhou, Yicong},
  booktitle={2020 IEEE International Conference on Multimedia and Expo (ICME)}, 
  title={Zero-Shot Restoration of Underexposed Images via Robust Retinex Decomposition}, 
  year={2020},
  volume={},
  number={},
  pages={1-6},
  doi={10.1109/ICME46284.2020.9102962}}

@article{HAI2023103712,
title = {R2RNet: Low-light image enhancement via Real-low to Real-normal Network},
journal = {Journal of Visual Communication and Image Representation},
volume = {90},
pages = {103712},
year = {2023},
issn = {1047-3203},
doi = {https://doi.org/10.1016/j.jvcir.2022.103712},
url = {https://www.sciencedirect.com/science/article/pii/S1047320322002322},
author = {Jiang Hai and Zhu Xuan and Ren Yang and Yutong Hao and Fengzhu Zou and Fang Lin and Songchen Han}
}

@ARTICLE{9328179,
  author={Yang, Wenhan and Wang, Wenjing and Huang, Haofeng and Wang, Shiqi and Liu, Jiaying},
  journal={IEEE Transactions on Image Processing}, 
  title={Sparse Gradient Regularized Deep Retinex Network for Robust Low-Light Image Enhancement}, 
  year={2021},
  volume={30},
  number={},
  pages={2072-2086},
  doi={10.1109/TIP.2021.3050850}}

@INPROCEEDINGS{9710730,
  author={Wang, Ruixing and Xu, Xiaogang and Fu, Chi-Wing and Lu, Jiangbo and Yu, Bei and Jia, Jiaya},
  booktitle={2021 IEEE/CVF International Conference on Computer Vision (ICCV)}, 
  title={Seeing Dynamic Scene in the Dark: A High-Quality Video Dataset with Mechatronic Alignment}, 
  year={2021},
  volume={},
  number={},
  pages={9680-9689},
  doi={10.1109/ICCV48922.2021.00956}}

@article{
doi:10.1126/science.1127647,
author = {G. E. Hinton  and R. R. Salakhutdinov },
title = {Reducing the Dimensionality of Data with Neural Networks},
journal = {Science},
volume = {313},
number = {5786},
pages = {504-507},
year = {2006},
doi = {10.1126/science.1127647},
URL = {https://www.science.org/doi/abs/10.1126/science.1127647},
eprint = {https://www.science.org/doi/pdf/10.1126/science.1127647}}

@misc{li2021allday, title={All-Day Object Tracking for Unmanned Aerial Vehicle}, author={Bowen Li and Changhong Fu and Fangqiang Ding and Junjie Ye and Fuling Lin}, year={2021}, eprint={2101.08446}, archivePrefix={arXiv}, primaryClass={cs.CV} }
}

\clearpage

\newpage
\twocolumn[%
  \begin{center}
    \Huge DRWKV: Focusing on Object Edges for Low-Light Image Enhancement\\
    \vspace{0.5em}
    \Large Appendices \& Supplementary Material
  \end{center}
  \vspace{1em}
]

\section{Supplementary Experiments}  
In this appendix, we first present a comprehensive exposition of the core computational pipeline of the DRWKV model introduced in the main text, together with the detailed implementation specifics of each key module. Subsequently, to rigorously validate the enhanced capabilities claimed by the proposed DRWKV model, we designed a series of extended experiments; the experimental protocol and the corresponding analyses of the results are elaborated in the ensuing sections.

\section{Architecture Details of DRWKV}
In this section, we provide a detailed exposition of the Light Preprocessing pipeline introduced in Section 3, the ES-RWKV computational flow, and the constituent fine-grained modules.
\subsection{Light Preprocessing}
Light Preprocessing takes a low-light image $I\in\mathbb{R}^{3\times H\times W}$ as input. First, under the gray-world hypothesis, it computes the global illumination component \(L_{\text{ill}}\in\mathbb{R}^{C\times H\times W}\). This hypothesis assumes that, in a color-balanced scene, the average reflectance of all pixels tends toward neutral gray. Guided by this assumption, the global illumination map \(L_{\text{ill}}\), which characterizes the spatial distribution of overall illumination intensity, is derived from the input image \(I\) and serves as the basis for the final enhancement.

The input image \(I\) is initially processed by a depthwise-separable \(3\times 3\) convolution layer with ReLU activation. The resulting feature map is then fed in parallel into three independent \(1\times 1\) convolution modules, each with a distinct activation function to produce specific components: a module with Sigmoid activation outputs the structured artifact component \(S\in\mathbb{R}^{3\times H\times W}\), intended to capture potential structured interference patterns; a module with Sigmoid activation outputs the noise estimation component \(N\in\mathbb{R}^{3\times H\times W}\), quantifying the noise contamination level; and a module with Tanh activation outputs the local illumination estimation component \(L\in\mathbb{R}^{1\times H\times W}\), providing a finer, spatially adaptive representation of illumination intensity. The procedure is as follows:

\begin{equation}
B(I)=\operatorname{ReLU}\!\bigl(\operatorname{DepthConv}_{3\times3}(I)\bigr),
\end{equation}
\begin{equation}
S=\operatorname{Sigmoid}\!\bigl(\operatorname{Conv}_{1\times1}(B^{3}(I))\bigr),
\end{equation}
\begin{equation}
L=\operatorname{Sigmoid}\!\bigl(\operatorname{Conv}_{1\times1}(B^{3}(I))\bigr),
\end{equation}
\begin{equation}
N=\operatorname{Tanh}\!\bigl(\operatorname{Conv}_{1\times1}(B^{3}(I))\bigr),
\end{equation}
where \(\operatorname{DepthConv}_{3\times3}(\cdot)\) denotes the \(3\times 3\) depthwise convolution, \(\operatorname{Conv}_{1\times1}(\cdot)\) denotes the \(1\times 1\) convolution, \(\operatorname{ReLU}(\cdot)\), \(\operatorname{Sigmoid}(\cdot)\), and \(\operatorname{Tanh}(\cdot)\) are the respective activation functions, and \(B^{3}(\cdot)\) denotes three consecutive applications of \(B(\cdot)\).

The noise estimation \(N\), illumination estimation \(L\), and the original input \(I\) are combined via
\begin{equation}
\hat{R}=\frac{I-N}{L},
\end{equation}
yielding the reconstructed reflectance \(\hat{R}\in\mathbb{R}^{3\times H\times W}\). Then, \(I\) and \(\hat{R}\) are combined to obtain the enhanced image \(\hat{I}\in\mathbb{R}^{3\times H\times W}\):
\begin{equation}
\hat{I}=I\odot\hat{R},
\end{equation}
where \(\odot\) denotes element-wise multiplication.
\begin{figure*}[t]
    \centering
    \includegraphics[width=\linewidth]{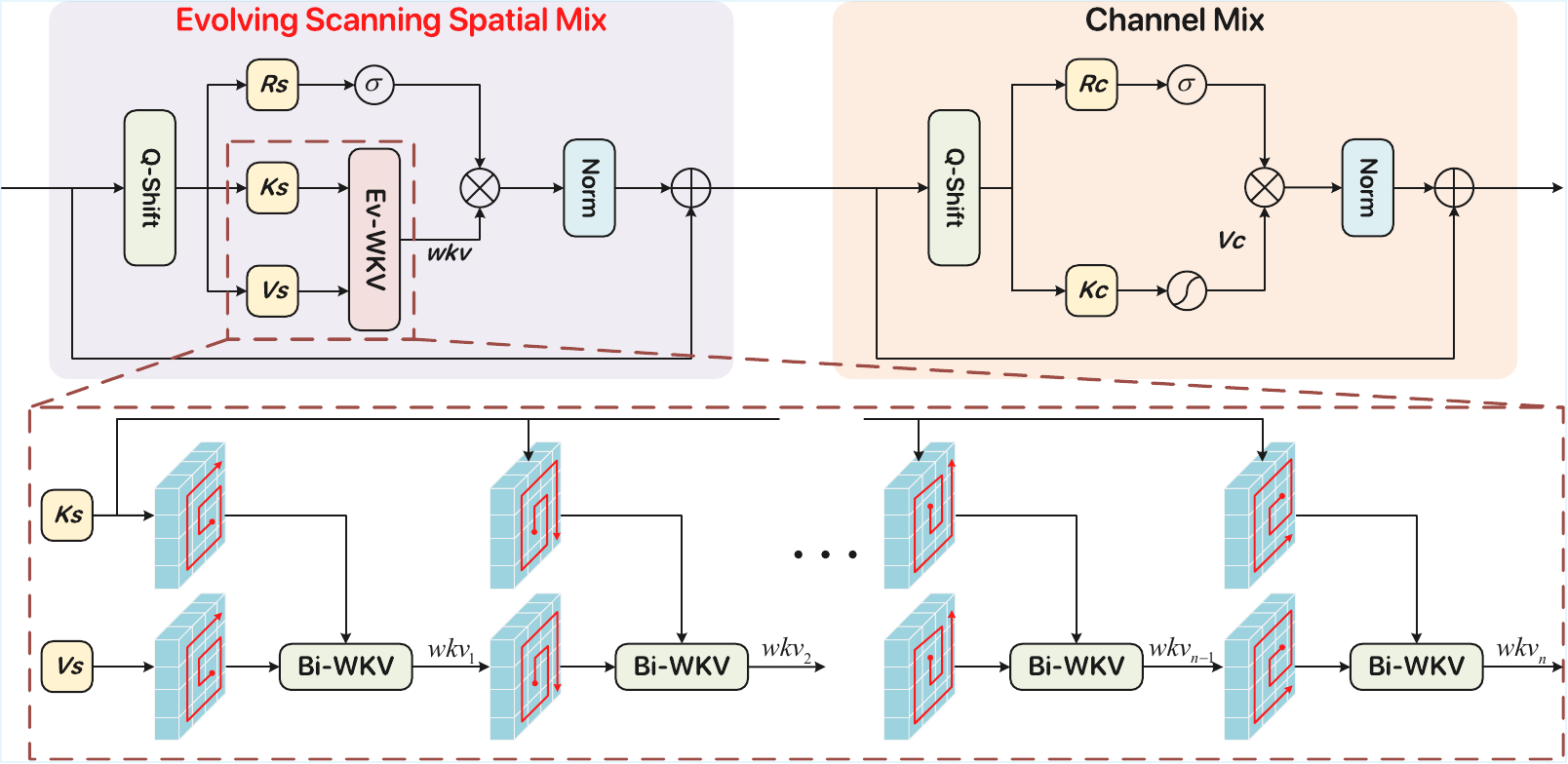}
    \caption{Evolving Scanning RWKV's structure, it consists of two parts: Evolving Scanning Spatial Mix and Channel Mix.}
    \label{fig:scanning}
\end{figure*}
\subsection{Evolving Scanning RWKV}
Unlike traditional spatial modelling approaches, the proposed ES-RWKV breaks through spatial-size constraints, preserves structural integrity, enhances modelling capacity, and enables cross-space information interaction while markedly reducing the computational overhead of handling complex spatial features, thereby offering an efficient solution for spatial feature extraction and fusion in vision tasks. The ES-RWKV block is decomposed into two sequential stages: Evolving Scanning Spatial Mix and Channel Mix. The Evolving Scanning RWKV structure is shown in Figure.\ref{fig:scanning}

\subsubsection{Evolving Scanning Spatial Mix}
They tackles the dilemma between spatial-size limits and structural integrity.  A channel-wise Q-Shift allows the model to adapt flexibly to different tasks without extra computation.  Coupled with the EV-WKV mechanism, an embedded eight-spiral scanning pattern is introduced to capture spatial dependencies more comprehensively and efficiently, suppressing redundancy irrelevant to spatial extraction.  Spatial information is then consolidated by a LayerNorm, enabling the model to automatically emphasize salient features and attenuate noise, thereby facilitating downstream layers in learning effective spatial patterns.

For an input feature $X \in \mathbb{R}^{T\times C}$, we augment the standard sequential processing framework with a channel-partitioned Q-Shift mechanism.By slicing and concatenating along the channel axis, we construct a shifted feature $X^{\dagger}$.Incorporating the learnable vector $\mu_{(\star)}$, we dynamically interpolate and fuse the original feature $X$ and the shifted feature $X^{\dagger}$ to further transcend the spatial size limitations of traditional methods, preserve the integrity of spatial structures, and enhance the model's capability to capture cross-spatial positional information. This process yields feature components tailored for spatial mixing computations:
\begin{equation}
\textit{Q-Shift}_{(\star)}(X)=X+(1-\mu_{(\star)})X^{\dagger}
\end{equation}
\begin{equation}
\begin{split}
X^{\dagger}[h,w]=\textit{Concat}\bigl(
  & X[h-1,w,0:C/4], \\
  & X[h+1,w,C/4:C/2], \\
  & X[h,w-1,C/2:3C/4], \\
  & X[h,w+1,3C/4:C]
\bigr)
\end{split}
\end{equation}

In this context, $(*) \in \{\textit{R},\textit{K},\textit{V}\}$ denotes three types of interpolation operations performed on $X$ and $X^{\dagger}$ . These operations are governed by $\mu$ and are utilized for the subsequent computation of \textit{R}, \textit{K}, and \textit{V}. Here, \textit{h}, \textit{w}, and \textit{C} represent the height, width, and number of channels, respectively. 

Subsequently, after the Q-Shift operation $X$ is decomposed into three independent components $R_s, K_s, V_s\in R^{T\times C}$
\begin{equation}
(*)_s=\textit{Q-Shift}_{(*)}(X)W_{(*)}=(X+(1-\mu_{(*)})X^{\dagger})W_{(*)}
\end{equation}

$W_{(*)}$ denotes the weight. To ensure numerical stability, prevent training divergence, and accelerate training convergence, we introduce the LN layer. The total output of the Evolving Scanning Spatial Mix block is as follows:
\begin{equation}
O_s=X+LN((\sigma (R_s)\odot wkv)W_{O_s}),
\end{equation}
\begin{equation}
\textit{where }wkv=\textit{EV-WKV}(K_s,V_s),
\end{equation}

Here, $\sigma$  denotes the sigmoid activation function, $\odot$ represents the element-wise multiplication operation, LN denotes layer normalization, and $W_{O_s}$ denotes the weight. EV-WKV is an Evolving Scanning Mechanism proposed based on the linear scanning formula wkvt of Bi-WKV.

\subsubsection{Channel Mix}
We reintroduce the Q-Shift mechanism and combine it with the SquaredReLU activation function and linear projection. This is aimed at addressing the shortcomings that may exist after the spatial mixing in the first stage, such as insufficient exploration of correlations between channel features and oversimplified feature expression. It ensures more adequate feature fusion in the channel dimension, enhances the nonlinear expression ability and discriminability of features, and further maintains the stability of numerical distribution through Layer Normalization (LN), thereby providing richer and more robust inputs for feature learning in the deep layers of the model.

First, we take the feature $Os$  output from the first stage as input, and construct the query feature $Rc$ and key feature $\textit{Kc}$ for channel interaction respectively through the Q-Shift mechanism. The calculation process is as follows:
\begin{equation}
\begin{split}
& R_c=Q-Shift_R(O_s)W_R,\\
& K_c=Q-Shift_K(O_s)W_K
\end{split}
\end{equation}

Here, $\textit{Q-Shift}_R$ and $\textit{Q-Shift}_K$ dynamically interpolate between the original feature $O_s$ and its shifted version $O^{\dagger}_s$ using learnable parameters $\mu_{(R)}$ and $\mu_{(K)}$. This design not only preserves the spatial structural information in the original features but also introduces cross-position channel dependencies, enabling the model to adaptively focus on long-range correlations among different channels.

Subsequently, to enhance the nonlinear expression capability of features and alleviate the gradient vanishing problem, the $\textit{SquaredReLU}$ activation function is applied to for $\textit{Kc}$ transformation.

Further, the transformed features are projected into the value space via linear projection, and the output of the Channel Mix module is generated through residual connections and layer normalization operations. The detailed process is as follows:
\begin{gather}
O_c=O_s+LN((\sigma (R_c)\odot V_c)W_{O_c}),\\
\textit{where }V_c=SquaredReLU(K_c)
\end{gather}

\subsection{Bilateral Spectrum Aligner Block}

This subsection focuses on the three core components of the Bilateral Spectrum Aligner: Spectral Alignment Enhancer (SAE), Cross Attention (CA), and Scharr Edge Enhancement (SEE).

\subsubsection{Spectral Alignment Enhancer (SAE)}
\begin{figure}[t]
    \centering
    \includegraphics[width=1\linewidth]{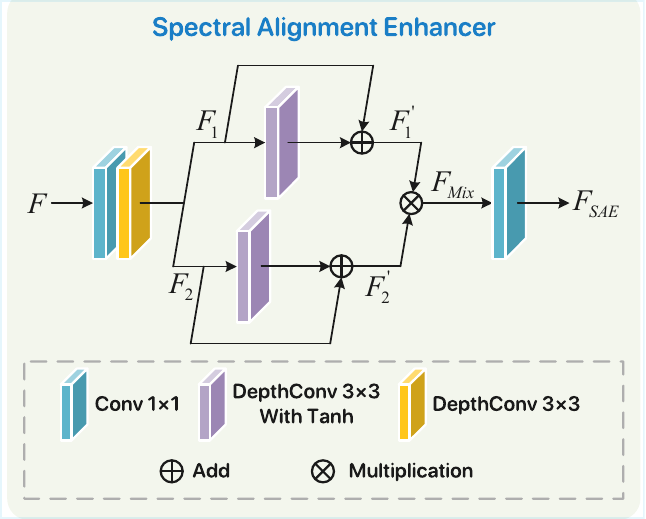}
    \caption{Spectral Alignment Enhancer's structure.}
    \label{fig:SAE}
\end{figure}
To enhance the low-light image enhancement model’s perception and processing of spectral features under varying illumination conditions, we propose an adaptive feature-enhancement module termed the Spectral Alignment Enhancer (SAE), illustrated in Figure~\ref{fig:SAE}. Specifically, the module achieves precise optimization of low-light features through three sequential stages: feature expansion, separated enhancement, and interactive fusion.

Initially, the input feature \(F\in\mathbb{R}^{C\times H\times W}\) is channel-wise expanded via a \(1\times1\) convolution and then split into two branches, thereby preserving sufficient expressive capacity for capturing faint spectral details in low-light environments.
\begin{equation}
F_1, F_2 = Split(DepthConv_{3\times3}(Conv_{1\times1}(F)))
\end{equation}

Here, $Split(\cdot)$ denotes an even channel-wise halving,.

Subsequently, the two feature streams \(F_1, F_2 \in \mathbb{R}^{C_1/2 \times H \times W}\) are each processed by a depthwise-separable convolution to focus on local spatial patterns, followed by a nonlinear Tanh activation and a residual connection, resulting in refined features \(F'_1, F'_2 \in \mathbb{R}^{C_1/2 \times H \times W}\). This design allows the SAE to enhance the extraction of critical low-light cues, such as edges and textures, while preventing detail loss caused by excessive transformations. The procedure can be expressed as:
\begin{equation}
F'_1=F_1+Tanh(DepthConv_{3\times3}(F_1))
\end{equation}
\begin{equation}
F'_2=F_2+Tanh(DepthConv_{3\times3}(F_2))
\end{equation}

Finally, the two refined feature streams are multiplied element-wise to achieve dynamic interaction, adaptively reinforcing informative cues. A subsequent \(1\times 1\) convolution then compresses the result back to the original channel dimension, yielding the optimized feature \(F_{\text{SAE}}\in\mathbb{R}^{C\times H\times W}\). This process is formulated as:
\begin{equation}
F_{SAE}=Conv_{1\times1}(F'_1\times F'_{2})
\end{equation}

\subsubsection{Cross Attention (CA)}
\begin{figure}[t]
    \centering
    \includegraphics[width=1\linewidth]{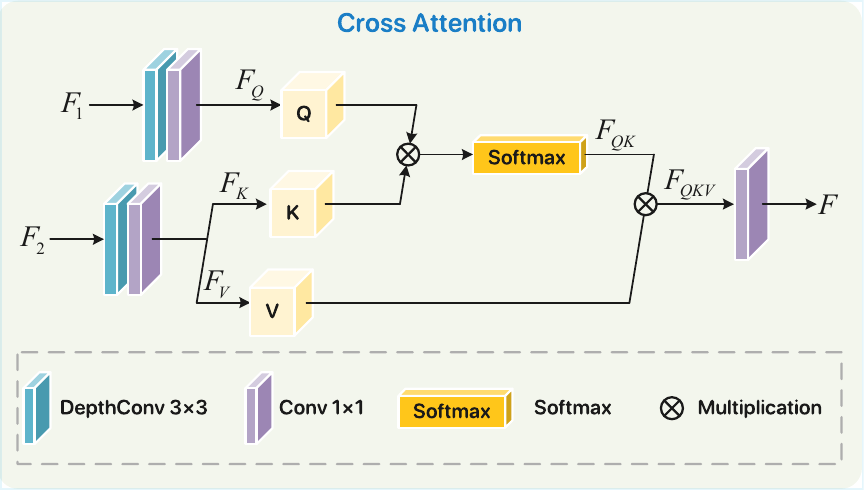}
    \caption{Cross Attention's structure.}
    \label{fig:CA}
\end{figure}
In low-light enhancement, the luminance stream x and the chrominance stream y are usually mis-aligned in both spatial statistics and semantic content.  We propose a lightweight Cross-Attention (CA) block that lets x query relevant information from y through a multi-head, L2-normalized cross-covariance attention implemented with depth-wise convolutions only.  The entire procedure is summarized below with strict correspondence to the released code, illustrated in Figure~\ref{fig:CA}.

First,
Given two feature maps:
\begin{equation}
    x,y\in R^{C\times H\times W}
\end{equation},
where x originates from the brightness subnet and y from the color subnet.

Query / Key / Value Generation

All linear projections are instantiated as 1×1 convolutions followed by 3×3 depth-wise convolutions to retain local context while being parameter-efficient.
\begin{equation}
    Q=DWConv_{3\times 3}(Conv_{1\times 1}(x))\in R^{C\times H\times W}
\end{equation}
\begin{equation}
    KV=DWConv_{3\times 3}(Conv_{1\times 1}(y))\in R^{2C\times H\times W}
\end{equation}
The KV tensor is then channel-split into key and value:
\begin{equation}
    K,V=split(KV,dim=1),K,V\in R^{C\times H\times W}
\end{equation}

Multi-Head Re-Shape and L2-Normalization

To exploit complementary attention patterns, we reshape each tensor into h heads:
\begin{equation}
    Q_{head}=rearrange(Q)\in R^{n_h\times c_h\times (H\cdot W)}
\end{equation}
\begin{equation}
    K_{head}=rearrange(K)\in R^{n_h\times c_h\times (H\cdot W)}
\end{equation}
\begin{equation}
    V_{head}=rearrange(V)\in R^{n_h\times c_h\times (H\cdot W)}
\end{equation}

To ensure training stability, L2-normalization is applied along the last dimension:
\begin{equation}
    \hat Q=\frac{Q_{head}}{\|Q_{head}\|_2},\hat K=\frac{K_{head}}{\|K_{head}\|_2}
\end{equation}
\begin{figure}[t]
    \centering
    \includegraphics[width=1\linewidth]{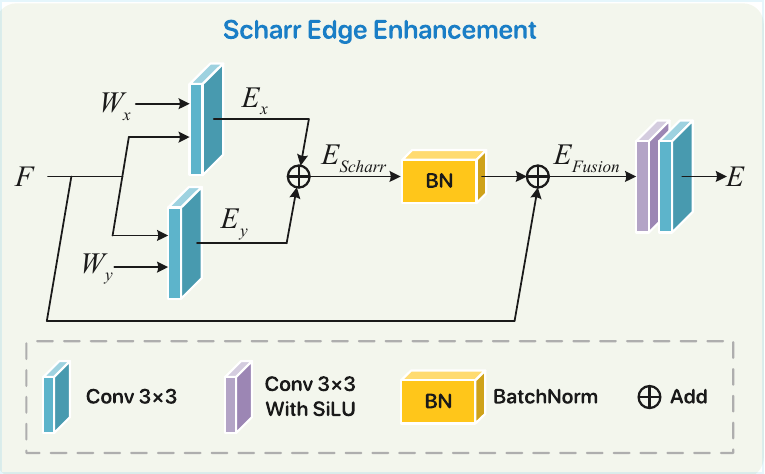}
    \caption{Scharr Edge Enhancement's structure.}
    \label{fig:SEE}
\end{figure}
Cross-Covariance Attention

We compute the cross-covariance attention matrix between x-query and y-key:
\begin{equation}
    attn=softmax(\frac{\hat Q\hat K^{T}}{\tau})\in R^{n_h\times c_h\times c_h}
\end{equation}
where the temperature \(\tau\) is a learnable scalar initialized to d
The attended feature is then aggregated via matrix multiplication with y-value:
\begin{equation}
   out=attn\cdot V_{head}\in R^{n_h\times c_h \times (H\cdot W)} 
\end{equation}

Output Projection

Finally, the multi-head output is re-arranged and projected back to the original channel dimension via a 1×1 convolution:
\begin{equation}
    Output = Conv_{1 \times 1}(rearrange(out))\in R^{C\times H\times W}
\end{equation}

The resulting tensor is element-wise added to the luminance branch, yielding an enhanced feature that is both color-aware and noise-suppressed.

\subsubsection{Scharr Edge Enhancement (SEE)}

In the field of low-light image enhancement, the accurate preservation of edge contours and the complete recovery of texture details have always been core challenges that restrict the performance of algorithms. In low-light environments, the image Signal-to-Noise Ratio (PSNR) decreases significantly, causing the gradient information in edge regions to be submerged by noise and thus become weak. Meanwhile, traditional enhancement methods, in the process of improving brightness, often further exacerbate edge blurring and texture loss due to excessive smoothing operations or noise amplification effects, ultimately leading to visual distortion in the enhanced results characterized by "improved brightness but lacking details", illustrated in Figure~\ref{fig:SEE}.
\subsection{User Subjective Evaluation Experiment}
\begin{figure*}[t]
    \centering
    \includegraphics[width=1\linewidth]{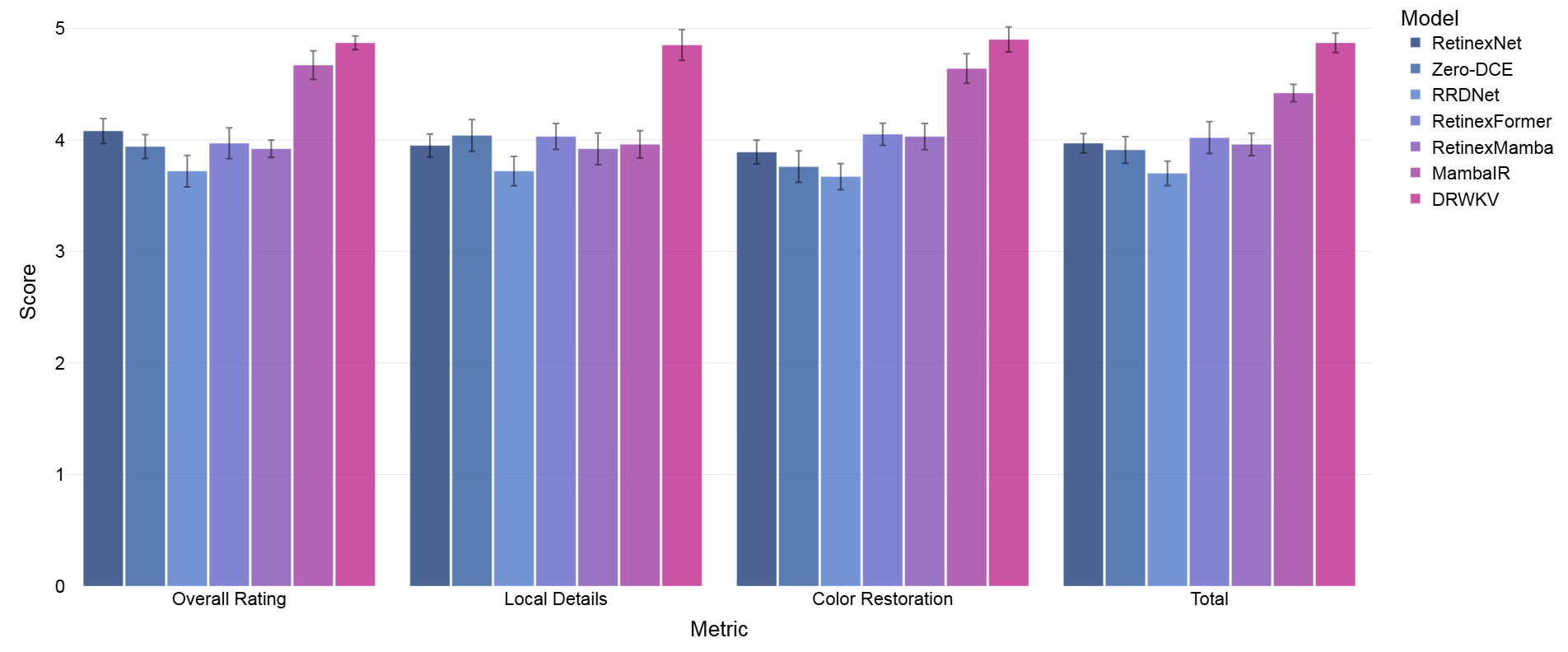}
    \caption{User subjective evaluation experiment.}
    \label{fig:4.1}
\end{figure*}

To address the above issues, we propose a Scharr Edge Enhancement (SEE) module. By integrating the prior knowledge of traditional image processing with the feature learning capability of deep learning, it provides an effective solution for structure-aware enhancement in low-light scenarios.
Specifically, given the input feature $F \in \mathbb{R}^{C \times H \times W}$, we first perform channel-wise filtering operations using pre-fixed Scharr convolution kernels and convolution weights $W_x$,$W_y$ to simulate the gradient extraction process in traditional image processing, obtaining the x-direction gradient feature $E_x \in \mathbb{R}^{C \times H \times W}$ and y-direction gradient feature $E_y \in \mathbb{R}^{C \times H \times W}$ respectively.

Subsequently, to suppress noise interference and aggregate multi-directional edge information, we fuse $E_x$ and $E_y$ using the L1 norm to compute the Scharr edge strength feature $E_{Scharr} \in \mathbb{R}^{C \times H \times W}$.

Then, the edge strength feature is integrated with the original input feature through a residual mechanism to obtain the fused feature $E_{Fusion} \in \mathbb{R}^{C \times H \times W}$. This operation, on one hand, preserves the structural prior in the original feature; on the other hand, it avoids "edge overexposure" caused by excessive feature transformation under the guidance of edge information, thereby ensuring the naturalness of edge enhancement.

Finally, to achieve adaptive adjustment of feature dimensions and deep fusion of semantic information, $1\times1$ convolution and $3\times3$ convolution are used for dimension reduction and recovery, resulting in the final output $SEE_{out} \in \mathbb{R}^{C \times H \times W}$.

The specific calculation process can be expressed as follows:
\begin{equation}
E_x = Conv_{3 \times 3}(F,W_x),E_y=Conv_{3 \times 3}(F,W_y)
\end{equation}
\begin{equation}
W_x = \begin{bmatrix} -3 & 0 & 3 \\ -10 & 0 & 10 \\ -3 & 0 & 3 \end{bmatrix},W_y = \begin{bmatrix} -3 & -10 & -3 \\ 0 & 0 & 0 \\ 3 & 10 & 3 \end{bmatrix}
\end{equation}
\begin{equation}
E_{Scharr} = \|E_x\|_1 + \|E_y\|_1
\end{equation}
\begin{equation}
E_{Fusion}=F+BN(E_{Scharr})
\end{equation}
\begin{equation}
SEE_{out} =Conv_{3\times3}(SiLU(Conv_{1\times1}(E_{Fusion})))
\end{equation}
Here, $Conv_{3\times3}(\cdot,\cdot)$ denotes a $3\times3$ convolution with fixed weights, $BN(\cdot)$ represents the BatchNorm operation, and $SiLU(\cdot)$ denotes the SiLU activation function.
\section{The input of the Global Edge Retinex theory}
In the main text, we elaborate and derive the proposed GER, and present the input structure in the main text figures: edge features $ E $, artifacts $ S $, and processed global illumination $ L_{ill} $. As the artifacts $ S $ and processed global illumination $ L_{ill} $ have been thoroughly explained, in the appendix, we discuss the extraction and integration of edge features $ E $.

To reduce the computational load and complexity of the model itself, we further utilize the SEE based on the output $I_{DRWKV}\in \mathbb{R}^{3 \times H \times W}$ of the DRWKV model, thereby obtaining the integration of input edge features $ E $ suitable for Global Edge Retinex theory.

The specific calculation process can be expressed as follows:
\begin{equation}
E = SEE(I_{DRWKV})
\end{equation}

\section{Extended experiment}
\subsection{User Subjective Evaluation Experiment}
To rigorously assess the perceptual quality of enhancement algorithms under complex real world conditions we designed and conducted a user subjective evaluation experiment. Ten low light images were randomly selected from the publicly available LSRW Huawei benchmark and processed by seven enhancement models: RetinexNet, Zero-DCE, RRDNet, RetinexFormer, RetinexMamba, MambaIR and the proposed DRWKV to generate corresponding enhanced results. One hundred participants spanning diverse ages and professional backgrounds were then recruited. In an independent double blind protocol participants assigned scores on a five point Likert scale from one worst to five best along three key perceptual dimensions overall visual quality local detail restoration and global color fidelity. The mean opinion scores are reported in Figure~\ref{fig:4.1}. Across all three dimensions the proposed DRWKV model achieved the highest ratings indicating superior perceptual enhancement performance from the human visual perspective.

\subsection{Low-light Object Tracking Experiment}
The aforementioned experiments have demonstrated that DRWKV already exhibits superior performance in basic experimental scenarios.  To investigate its generalization ability, we designed a low-light multi-object tracking task.  For the dataset, we adopted the Drone low-light tracking benchmark, UAVDark135.  It contains 135 video sequences in total, encompassing various tracking scenarios such as intersections and highways, as well as tracking targets including pedestrians, boats, and vehicles.  The videos are captured at a frame rate of 30 FPS with a resolution of 1920×1080.

In terms of methodology, we compared the performance of mainstream object tracking algorithms with and without the integration of the DRWKV low-light enhancement module, using MOTA (multi-object tracking accuracy), IDF1 (identity recognition F1 score), and HOTA (high-order tracking accuracy) as evaluation metrics.

The experimental conclusions show that DRWKV generally improved accuracy across multiple mainstream tracking algorithms, with particularly prominent performance in multi-object tracking and identity preservation.  Meanwhile, the limitations of this module in optimizing Deep OC-SORT warn us that in the face of low adaptability and weak tuning of network structures, it may not be possible to achieve simultaneous improvements in detection accuracy and association capability.

\subsection{$\alpha$, $\beta$ and $\gamma$ Parameter Detail Experiment}
To obtain the optimal configuration of the three parameters ($\alpha$, $\beta$ and $\gamma$), we designed a series of parameter detail experiments. For the dataset, we adopted the LOLv2-Real dataset. During the implementation of the parameter detail experiments, we only adjusted one of the three parameters ($\alpha$, $\beta$ and $\gamma$) while keeping the other two unchanged. The experimental results are presented in Table~\ref{tab:values1}, Table~\ref{tab:values2}, and Table~\ref{tab:values3}. These results indicate that the optimal parameter configuration for DRWKV is $\alpha=0.2$, $\beta=-0.05$, and $\gamma=0.2$.
\begin{table}[htbp]
    \centering
    \fontsize{8.2pt}{10.2pt}\selectfont 
    \setlength{\tabcolsep}{5pt}     
    \renewcommand{\arraystretch}{1.2} 
    \caption{The impact of different $\alpha$ values on the performance of the LOLv2-Real dataset.}
    \label{tab:values1}
    \begin{tabular}{lccccccc}
        \toprule
        \quad$\alpha$ & 0.05 & 0.10 & 0.15 & \textbf{0.20} & 0.25 & 0.30 & 0.35\\
        \midrule
        PSNR↑ & 23.56 & 23.89 & 24.02 & \textbf{24.12} & 23.84 & 23.97 & 23.94\\
        SSIM↑ & 0.798 & 0.815 & 0.814 & \textbf{0.832} & 0.792 & 0.821 & 0.814\\
        NIQE↓ & 4.012 & 3.967 & 3.941 & \textbf{3.926} & 3.951 & 3.989 & 4.173\\
        \bottomrule
    \end{tabular}
\end{table}
\begin{table}[htbp]
    \centering
    \fontsize{8.2pt}{10.2pt}\selectfont 
    \setlength{\tabcolsep}{8pt}     
    \renewcommand{\arraystretch}{1.2} 
    \caption{The impact of different $\beta$ values on the performance of the LOLv2-Real dataset.}
    \label{tab:values2}
    \begin{tabular}{lccccc}
        \toprule
        \quad$\beta$ & -0.100 & -0.075 & \textbf{-0.050} & -0.025 & 0 \\
        \midrule
        PSNR↑ & 23.72 & 23.98 & \textbf{24.12} & 23.86 & 23.61\\
        SSIM↑ & 0.803 & 0.821 & \textbf{0.832} & 0.817 & 0.804\\
        NIQE↓ & 3.995 & 3.948 & \textbf{3.926} & 3.973 & 4.003\\
        \bottomrule
    \end{tabular}
\end{table}
\begin{table}[htbp]
    \centering
    \fontsize{8.2pt}{10.2pt}\selectfont 
    \setlength{\tabcolsep}{5pt}     
    \renewcommand{\arraystretch}{1.2} 
    \caption{The impact of different $\gamma$ values on the performance of the LOLv2-Real dataset.}
    \label{tab:values3}
    \begin{tabular}{lccccccc}
        \toprule
        \quad$\gamma$ & 0.05 & 0.10 & 0.15 & \textbf{0.20} & 0.25 & 0.30 & 0.35\\
        \midrule
        PSNR↑ & 22.74 & 23.65 & 23.91 & \textbf{24.12} & 23.99 & 23.78 & 23.84\\
        SSIM↑ & 0.804 & 0.801 & 0.823 & \textbf{0.832} & 0.826 & 0.812 & 0.814\\
        NIQE↓ & 4.112 & 4.023 & 3.957 & \textbf{3.926} & 3.962 & 3.998 & 4.472\\
        \bottomrule
    \end{tabular}
\end{table}

\end{document}